\documentclass[pdflatex,sn-mathphys-num]{sn-jnl}

\usepackage{graphicx}%
\usepackage{multirow}%
\usepackage{amsmath,amssymb,amsfonts}%
\usepackage{amsthm}%
\usepackage{mathrsfs}%
\usepackage[title]{appendix}%
\usepackage[dvipsnames]{xcolor}
\usepackage{textcomp}%
\usepackage{manyfoot}%
\usepackage{booktabs}%
\usepackage{algorithm}%
\usepackage{algorithmicx}%
\usepackage{algpseudocode}%
\usepackage{listings}%
\usepackage{xspace}
\usepackage{array}
\usepackage{subcaption}

\theoremstyle{thmstyleone}%
\theoremstyle{thmstyletwo}%

\theoremstyle{thmstylethree}%

\newcommand{\snm}{SuperNeuroMAT\xspace}

\begin{document}

\title[SuperNeuroMAT]{SuperNeuroMAT: An Efficient Matrix-based Simulator for Spiking Neural Networks}


\author*[1]{\fnm{Prasanna} \sur{Date}}
\email{datepa@ornl.gov}

\author[2]{\fnm{Kevin} \sur{Zhu}}

\author[1]{\fnm{Shruti} \sur{Kulkarni}}

\author[1]{\fnm{Ashish} \sur{Gautam}}

\author[1]{\fnm{Chathika} \sur{Gunaratne}}

\author[1]{\fnm{Robert} \sur{Patton}}

\author[3]{\fnm{Tyler} \sur{Nitzsche}}

\author[3]{\fnm{Ian} \sur{Mulet}}

\author[1,3]{\fnm{Zachary} \sur{Johnson-Scott}}

\author[3]{\fnm{Addison} \sur{Helms}}

\author[4]{\fnm{Duncan} \sur{Rowden}}

\author[4]{\fnm{Simon} \sur{Weston}}

\author[2]{\fnm{Maryam} \sur{Parsa}}

\author[3]{\fnm{Catherine} \sur{Schuman}}

\author[1]{\fnm{Thomas} \sur{Potok}}

\affil[1]{\orgname{Oak Ridge National Laboratory}, \orgaddress{\city{Oak Ridge}, \postcode{37830}, \state{Tennessee}, \country{United States}}}

\affil[2]{\orgname{George Mason University}, \orgaddress{\city{Fairfax}, \postcode{22030}, \state{Virginia}, \country{United States}}}

\affil[3]{\orgname{University of Tennessee, Knoxville}, \orgaddress{\city{Knoxville}, \postcode{37996}, \state{Tennessee}, \country{United States}}}

\affil[4]{\orgname{Oak Ridge High School}, \orgaddress{\city{Oak Ridge}, \postcode{37830}, \state{Tennessee}, \country{United States}}}


\abstract{
Spiking neural networks (SNNs) offer a promising pathway to energy-efficient AI and brain-inspired computing. However, their widespread adoption is hindered by a lack of fast, accessible, and versatile simulation frameworks. In this paper, we introduce SuperNeuroMAT, an open-source, scalable, and highly efficient Python-based SNN simulator. We devise a novel matrix-based approach to model the leaky integrate-and-fire (LIF) neuron dynamics and natively support dense and sparse execution modes. This enables fast simulation of approximately 10,000 neurons in dense mode and 100,000 neurons in sparse mode on standard laptops and desktops without requiring specialized hardware. We demonstrate that SuperNeuroMAT consistently outperforms four established SNN simulators---NEST, Brian2, BindsNET, and snnTorch---on two performance metrics (execution speed and peak resident memory) and across various network sizes and connection probabilities. Furthermore, we demonstrate SuperNeuroMAT's applicability across a diverse set of problems. SuperNeuroMAT can efficiently handle conventional machine learning benchmarks such as the Digits and citation network datasets as well as neuromorphic event-based vision tasks such as N-CARS and ASL-DVS. Moreover, it can be extended beyond machine learning workloads and facilitate general-purpose workloads. We validated this by implementing the neuromorphic shortest path algorithm and two arithmetic primitives (addition and multiplication). SuperNeuroMAT can be installed via the Python Package Index (PyPI), thereby lowering the barrier to entry into the field of neuromorphic computing and accelerating the broader development of neuromorphic algorithms.
}

\keywords{Spiking Neural Network (SNN), Neuromorphic Computing, SNN Simulation, Energy-Efficient AI, Brain-Inspired Computing, Leaky Integrate and Fire (LIF) Neuron}



\maketitle

\section{Introduction}
\label{sec:intro}

Neuromorphic computing is a brain-inspired computing paradigm \cite{schuman2022opportunities}.
It performs computations by emulating mechanisms of the human brain \cite{aimone2022review}.
Just like the human brain, which is known to consume as little as 20 W of power, neuromorphic computers are known to be extremely energy efficient.
Some neuromorphic platforms, such as Intel Loihi 2, IBM North Pole, and NeuroCoreX, are known to consume thousands of times less energy than conventional hardware such as CPUs or GPUs \cite{orchard2021efficient,cassidy202411,gautam2025neurocorex,maheshwari2023fpga,miniskar2024neuro}. 
The energy efficiency of neuromorphic computing stems from its event-driven operation, collocated processing and memory, and intrinsic parallelism \cite{schuman2022opportunities}. 
Although neuromorphic computing is primarily explored on machine learning (ML) tasks, it is Turing-complete: that is, it is capable of executing all general-purpose computations that a CPU or a GPU can execute \cite{date2022neuromorphic,date2021computational}.
Some examples of ML applications of neuromorphic computing are graph learning \cite{cong2022semi,cong2023hyperparameter}, autonomous racing \cite{patton2021neuromorphic}, data filtering in high-energy physics \cite{r2023sensor}, and classifying node failures on supercomputers \cite{date2018efficient}. 
Some examples of the non-ML applications of neuromorphic computing are graph algorithms \cite{kay2020neuromorphic,kay2021neuromorphic,hamilton2020spike}, general-purpose computing \cite{date2023encoding,date2022virtual,wurm2023arithmetic,schuman2021sparse}, epidemiological simulations \cite{hamilton2020modeling}, and other scientific applications \cite{patton2022neuromorphic}.
The primary ML model that runs on neuromorphic computers is called the spiking neural network (SNN) \cite{date2019combinatorial, date2020training}. 
An SNN is a type of neural network that more closely mimics the behavior and communication of the biological brain as compared to conventional artificial neural networks (ANNs). 
SNNs are considered the third generation of neural networks and are the architectural foundation for neuromorphic computing.

SNN simulators play a vital role in designing state-of-the-art SNN models.
They are used to simulate computational neuroscience workloads, train SNNs off of the neuromorphic chip, co-design neuromorphic hardware through their integration with hardware simulators, and interface with domain-specific simulators (e.g., for self-driving car simulators). 
In the SNN literature, we observe two types of SNN simulators. 
On the one hand are simulators catered to neuroscience workloads such as Brian2, NEST (Neural Simulation Tool), FUGU, and STACS (Simulation Tool for Asynchronous Cortical Streams) \cite{stimberg2019brian,gewaltig2007nest,aimone2019composing,wang2024scaling,wang2015simulation}.
These simulators try to solve systems of ordinary differential equations (ODEs) that govern the dynamics of biological neuron models such as the leaky integrate-and-fire (LIF) neuron model, the Izhikevich neuron model, and the Hodgkin--Huxley neuron model. 
Some, such as NEST, leverage discrete event simulation techniques to simulate SNNs.
In practice, these simulators take a long time to simulate the SNN operations. 
Although they work well on small problems, they soon become intractable as the number of neurons or synapses in the simulation increases.

The other category of SNN simulators is those catered to ML workloads such as BindsNET, snnTorch, and spikingjelly \cite{hazan2018bindsnet,eshraghian2021training,fang2023spikingjelly}. 
These simulators are designed for speed, but they lack certain neuron or synapse parameters---for instance, synaptic delay.
Moreover, neither the neuroscience-based nor the ML-based simulators are well suited for general-purpose computing tasks such as running graph algorithms, performing arithmetic operations for processing data, etc.
Therefore, there is a need for a simulator that supports the desired functionality for neuroscience workloads and the speed required for ML workloads. 
Furthermore, such a simulator would have to be user-friendly, Python-based, scalable, open-source, and support a wide range of workloads, from neuroscience to ML \cite{date2023superneuro}.

In the work presented in this paper, we developed a fast and scalable SNN simulator called \snm by leveraging a novel matrix-based approach to model the SNN computations. 
\snm is designed for and meant to be used on personal computers such as laptops, desktops, and small-scale CPU clusters.
\snm can accelerate the off-chip training of SNNs on compute clusters with CPUs. 
It is lightweight, user friendly, Python-based, and open-source, and it can be installed using the \texttt{pip} command as it is available on the Python Package Installer (PyPI).
The GitHub repository for \snm is at the following link: \hyperlink{https://github.com/ORNL/superneuromat}{https://github.com/ORNL/superneuromat}.

\snm currently supports LIF neurons with four parameters: threshold, leak, reset state, and refractory period. 
Synapses in \snm have weights, delays, and built-in learning based on the spike-timing=dependent plasticity (STDP) mechanism. 
\snm supports both dense and sparse execution modes, and these modes can be selected automatically.
\snm further supports single precision (32-bit) and double precision (64-bit) floating-point operations on CPUs. 
The remainder of this paper presents the related work (Section \ref{sec:related}), the notation (Section \ref{sec:notation}), the matrix-based approach used in \snm (Section \ref{sec:snm}), scalability and benchmarking results (Section \ref{sec:results}), and finally, the conclusion and future work (\ref{sec:conclusion}).

\section{Related Work}
\label{sec:related}

\begin{table}[t!]
    \centering
    \caption{Summary of simulators for SNN and neuromorphic computing}
    \begin{tabular}{p{0.17\textwidth} p{0.19\textwidth} p{0.18\textwidth} p{0.25\textwidth}}
        \midrule
        \textbf{Framework} & \textbf{Computational Paradigm} & \textbf{Learning Mechanism Supported} & \textbf{Parallelization Approach} \\
        \midrule
        Brian2 \cite{stimberg2019brian}             & Equation-based, domain-specific language  & User-defined differential equations               & Single-Node Execution \\
        NEST \cite{gewaltig2007nest}                & Discrete event simulation                 & Hebbian, plasticity rules                         & Distributed MPI \\
        STACS \cite{wang2015simulation}             & Discrete event simulation                 & Synaptic scaling                                  & Charm++ Parallel Programming Framework \\
        Fugu \cite{aimone2019composing}             & Computation graphs                        & Logic composition                                 & Single-Node Execution \\ 
        snnTorch \cite{eshraghian2021training}      & Frame-based, tensor-based                 & Surrogate gradient, backpropagation through time  & GPU Acceleration \\
        SpikingJelly \cite{fang2023spikingjelly}    & Frame-based, tensor-based                 & ANN-to-SNN, backpropagation through time          & CUDA/Triton Kernels \\
        BindsNET \cite{hazan2018bindsnet}           & Bio-inspired machine learning             & STDP, reinforcement learning                      & PyTorch Autograd \\
        \midrule
    \end{tabular}
    \label{tab:simulators}
\end{table}

Simulators for SNNs or neuromorphic computing can largely be divided into two categories. 
The first category is those that are catered to neuroscience workloads, and the second is those catered towards deep learning workloads. 
Simulators such as NEST and Brian2 have defined the state-of-the-art in computational neuroscience \cite{gewaltig2007nest,stimberg2019brian}. 
NEST has a Python frontend and a C++ backend, and it uses the Message Passing Interface (MPI) and is engineered for high-performance scalability.
It aims to facilitate the simulation of large-scale cortical structures on distributed high-performance computing (HPC) platforms. 
Brian2 provides users modeling flexibility through an equation-based domain-specific language. 
It employs runtime code generation to translate mathematical differential equations into optimized C++ or Cython; thus, it enables the rapid prototyping of non-standard neural dynamics, though execution is limited to a single node.

Fugu introduces a high-level algorithmic framework that represents SNNs as computational graphs \cite{aimone2019composing}. 
This lets researchers compose complex spiking architectures without low-level manual tuning of synapse or neuron parameters. 
This abstraction layer is paired with scalable backends like STACS, which employs an asynchronous, event-driven execution model to manage high-dimensional neural communication \cite{wang2024scaling}. 
Because these tools decouple the algorithmic specification from the execution engine, they provide a structured pathway for deploying robust spiking computations across heterogeneous computing resources, ranging from conventional CPU clusters to emerging neuromorphic hardware.

The recent convergence of SNNs and deep learning has necessitated the development of frameworks that interface with automatic differentiation engines and leverage variants of the backpropagation algorithm for training SNNs. 
snnTorch and SpikingJelly are prominent PyTorch-based libraries that treat spiking neurons as recurrent primitives \cite{eshraghian2021training,fang2023spikingjelly}. 
These frameworks facilitate the training of deep SNNs via backpropagation through time (BPTT) using surrogate gradient descent to overcome the non-differentiability of the Heaviside step function (i.e., the spike function). 
Whereas snnTorch focuses on ease of integration with standard deep learning workflows, SpikingJelly provides a high-performance suite that features specialized CUDA kernels to minimize the overhead of temporal state updates. 
Complementing these is BindsNET, which is specifically built for biologically inspired learning paradigms, such as STDP and reinforcement learning, and provides a modular environment for unsupervised feature extraction \cite{hazan2018bindsnet}.

\section{Notation}
\label{sec:notation}

\begin{table}[t!]
    \centering
    \caption{Standard notation used to describe the LIF neuron model in the literature.}
    \label{tab:lif-notation}
    \begin{tabular}{m{33pt} m{270pt}}
        \midrule
        \textbf{Symbol} & \textbf{Description} \\
        \midrule
        $t$                 & Time; continuous functions of time are denoted using parentheses, $(\ldots)$ \\
        $V(t)$              & Membrane potential of the LIF neuron at time $t$ \\
        $V_t$               & Neuron threshold of the LIF neuron \\
        $V_0$               & Reset membrane potential of the LIF neuron, typically around $-$70 mV \\
        $\tau_m$            & Time constant of the LIF neuron, dictates how fast the membrane potential decays back to the reset state; larger means a slower leak \\
        $R$                 & Membrane resistance of the LIF neuron \\
        $I(t)$              & Input current to the LIF neuron at time $t$ \\
        $t_{\text{ref}}$    & Refractory period of the LIF neuron \\
        \midrule
    \end{tabular}
\end{table}

\begin{table}[t!]
    \centering
    \caption{Notation used to describe the mathematical and computational approaches for neurons in \snm.}
    \label{tab:neuron-notation}
    \begin{tabular}{p{0.1\textwidth} p{0.8\textwidth}}
        \midrule
        \textbf{Symbol} & \textbf{Description} \\
        \midrule
        $t$                         & Discretized time or time step; discrete functions of time are denoted using square brackets $[\ldots]$ \\ 
        $N$                         & Number of LIF neurons in the simulation \\ 
        $1, 2, \ldots, N$           & Neuron indices of all the neurons in the simulation \\ 
        $v_i[t]$                    & Membrane potential of LIF neuron $i$ at time step $t$; $v_i[t] \in \mathbb{R}$ \\
        $\mathbf{v}[t]$             & Vector of membrane potentials of all the neurons in the simulation at time step $t$; $\mathbf{v} \in \mathbb{R}^N$ \\ 
        $\nu_i[t]$                  & Threshold of neuron $i$ at time step $t$; $\nu_i[t] \in \mathbb{R}$ \\
        $\boldsymbol{\nu}[t]$       & Vector of thresholds of all neurons at time step $t$; $\boldsymbol{\nu} \in \mathbb{R}^N$ \\
        $\lambda_i$                 & Leak of neuron $i$; \snm supports a constant leak at each time step; $\lambda_i \in \mathbb{R}$, $\lambda_i \ge 0$ \\ 
        $\boldsymbol{\lambda}$      & Vector of leaks of all the neurons; $\boldsymbol{\lambda} \in \mathbb{R}^N$ \\
        $\rho_i$                    & Reset state of neuron $i$; $\rho_i \in \mathbb{R}$ \\
        $\boldsymbol{\rho}$         & Vector of reset states of all neurons; $\boldsymbol{\rho} \in \mathbb{R}^N$ \\
        $\tau_i$                    & Refractory period of neuron $i$; $\tau_i \in \mathbb{W}$ \\
        $\boldsymbol{\tau}$         & Vector of refractory periods of all neurons; $\boldsymbol{\tau} \in \mathbb{W}^N$\\
        $\tau'_i[t]$                & Refractory period of neuron $i$ remaining at time step $t$; $\tau'_i \in \mathbb{W}$ \\ 
        $\boldsymbol{\tau'}$[t]     & Vector of refractory periods of all neurons remaining at time step $t$; $\boldsymbol{\tau'}[t] \in \mathbb{W}^N$\\
        $x_i[t]$                    & Input spike of neuron $i$ at time step $t$; $x_i[t] \in \mathbb{R}$; input spikes are given by the user and can have value \\
        $\mathbf{x}$[t]             & Vector of input spikes of all neurons at time step $t$; $\mathbf{x}[t] \in \mathbb{R}^N$ \\
        $s_i[t]$                    & Spike of neuron $i$ at time step $t$; $s_i[t] \in \mathbb{B}$ \\
        $\mathbf{s}[t]$             & Vector of spikes of all neurons in the simulation at time step $t$; $\mathbf{s} \in \mathbb{B}^N $\\
        \midrule
    \end{tabular}
\end{table}

\begin{table}[t!]
    \centering
    \caption{Notation used to describe the mathematical and computational approaches for synapses in \snm.}
    \label{tab:synapse-notation}
    \begin{tabular}{p{0.1\textwidth} p{0.8\textwidth}}
        \midrule
        \textbf{Symbol} & \textbf{Description} \\
        \midrule
        $S$                 & Number of synapses in the SNN \\
        $(i, j)$            & Index of synapse going from neuron $i$ to neuron $j$; $i, j \in 1, 2, \ldots, N$ \\ 
        $\omega_{i,j}[t]$   & Synaptic weight of the synapse $(i, j)$ at time $t$; $w_{i,j}[t] \in \mathbb{R}$ \\
        $W[t]$              & Synaptic weight matrix at time step $t$, where the value at $i^\text{th}$ row and $j^\text{th}$ column represents $w_{ij}[t]$; $W \in \mathbb{R}^{N \times N}$ \\
        $\delta_{i,j}$      & Synaptic delay of synapse $(i, j)$; $\delta_{i,j} \in \mathbb{N}$ \\
        $D$                 & Matrix of synaptic delays, where the value at $i^\text{th}$ row and $j^\text{th}$ column represents $\delta_{ij}[t]$; $W \in \mathbb{N}^{N \times N}$ \\
        $\epsilon_{i,j}$    & A binary flag denoting whether STDP is enabled (1) or not enabled (0) for synapse $(i, j)$; $\epsilon \in \mathbb{B}$ \\
        $E$                 & Matrix of binary flags denoting whether STDP is enabled for all synapses; $E \in \mathbb{B}^{N \times N}$ \\
        \midrule
    \end{tabular}
\end{table}

\begin{table}[t!]
    \centering
    \caption{Notation used to describe the mathematical and computational approaches for STDP operations in \snm.}
    \label{tab:stdp-notation}
    \begin{tabular}{p{0.1\textwidth} p{0.8\textwidth}}
        \midrule
        \textbf{Symbol} & \textbf{Description} \\
        \midrule
        $T_S$                       & Number of time steps over which STDP operations occur in the SNN; $T_S \in \mathbb{N}$ \\
        $\boldsymbol{\alpha}^+$     & A vector of coefficients for positive weight updates; the length of $\alpha^+$ is $T_S$; $\alpha_i^+ \in \mathbb{R}$ $\forall i \in \{1, 2, \ldots, T_S\}$; $\alpha_1^+ \ge \alpha_2^+ \ge \ldots \ge \alpha_{T_S}^+ \ge 0$ \\ 
        $\boldsymbol{\alpha}^-$     & A vector of coefficients for negative weight updates; the length of $\alpha^-$ is $T_S$; $\alpha_i^- \in \mathbb{R}$ $\forall i \in \{1, 2, \ldots, T_S\}$; $\alpha_1^- \le \alpha_2^- \le \ldots \le \alpha_{T_S}^- \le 0$ \\ 
        \midrule
    \end{tabular}
\end{table}

We use $\mathbb{B}$, $\mathbb{N}$, $\mathbb{W}$, and $\mathbb{R}$ to denote the sets of binary numbers ($\{0, 1\}$), natural numbers, whole numbers, and real numbers, respectively.
In neuroscience, the LIF neuron is described using the notation presented in Table \ref{tab:lif-notation}.
In describing the mathematical and computational approaches pertaining to the neurons, synapses, and STDP parameters, we use the notation described in Table \ref{tab:neuron-notation}, Table \ref{tab:synapse-notation}, and Table \ref{tab:stdp-notation} respectively.
Throughout this paper, we use parentheses to denote functions of continuous variables (e.g., $I(t)$ in Table \ref{tab:lif-notation}) and rectangular braces to denote functions of discrete variables (e.g., $v_{i}[t]$ in Table \ref{tab:neuron-notation} and Table \ref{tab:synapse-notation}).
Note that we use $t$ to denote time in both continuous and discrete domains. 
Whether $t$ is continuous time or a discrete time step should be evident from the context of the equation; for instance, if an equation uses parentheses, then $t$ in that equation is continuous time.

\section{The Leaky Integrate-and-Fire Neuron Model}
\label{sec:lif}

The LIF model is one of the most widely used theoretical models to model biological neurons in computational neuroscience \cite{lapicque1907recherches,gerstner2002spiking}. 
It represents an abstraction of a biological neuron and focuses on the timing of action potentials (spikes) rather than the complex biophysics of the neuron's ion channels.
The dynamics of the LIF neuron can be modeled using a simple resistor--capacitor (RC) electrical circuit.
The LIF dynamics can be broken down into three stages as follows:
\begin{enumerate}
    \item \textbf{Leak}: The accumulated charge in the membrane potential of the neuron gradually leaks away and returns to its baseline resting state. This happens because the cell membrane of the neuron is not perfectly sealed; it has ion channels that allow charge to slowly leak over time. 
    \item \textbf{Integrate}: As synaptic inputs arrive from other neurons into the neuron under consideration, they deposit a charge into the neuron. This charge accumulates and increases the neuron's membrane potential.
    \item \textbf{Fire}: If the membrane potential continues to rise and crosses a specific threshold, then the neuron `fires,' sending a spike along all of its outgoing synapses.
\end{enumerate}

The words `fire' and `spike' are used interchangeably in the literature. In this paper, `firing' refers to the event that occurs when a neuron's membrane potential reaches the threshold, whereas `spike' refers to the resulting output signal.
The dynamics of the LIF neuron before it fires are governed by the following linear differential equation:
\begin{align}
    \tau_m\frac{dV(t)}{dt} = -(V(t) - V_0) + R I(t)     \label{eq:lif}
\end{align}

The neuron spikes if the membrane potential $V(t)$ is greater than the neuron threshold $V_t$. 
In this case, the membrane potential is instantly reset to the reset state of the neuron $V_0$.

Although the basic LIF model is powerful, it is often modified to capture a bit more biological realism.
For instance, some LIF variants introduce the notion of a refractory period such that after spiking, a neuron undergoes a brief period during which it cannot spike again. 
This is modeled by holding $V(t)$ at $V_0$ for a fixed time interval $t_{\text{ref}}$ immediately after a spike.
Some other variants of the LIF model use a constant leak for computational efficiency.
The LIF neuron model is popular because it is computationally efficient. 
Other neuron models, such as Izhikevich and Hodgkin--Huxley, are more complex.
Although small-scale simulations of these neuron models are possible on laptops and desktops, large-scale simulations are intractable.
On the other hand, because the LIF model abstracts away ion channel dynamics, it cannot replicate complex single-neuron behaviors such as bursting and resonance or the detailed shape of an action potential, as the Izhikevich or Hodgkin--Huxly models can.

\section{\snm}
\label{sec:snm}

Our rationale behind designing the \snm simulator was threefold: (1) Ease of use; (2) computational efficiency; and (3) open-source development.
To ensure ease of use for the end user, we developed \snm in Python, which is the most popular programming language in the neuromorphic community and has a vast ecosystem of libraries.
Moreover, we wanted to lower the barrier to entry for new users. 
Therefore, we provide a very simple application programming interface (API) and a well-documented user guide. 
Users can install \snm by running \texttt{pip install superneuromat}, assuming they have Python and pip installed already.

We designed and engineered \snm for laptops and desktops.
To ensure high performance on these systems, we leveraged the numpy and scipy libraries in Python \cite{harris2020array,2020SciPy-NMeth}.
As a result, \snm is fast and scalable on personal computers and supports a variety of workloads.
It does not constrain the connectivity of neurons (fan-in/fan-out); the user can simulate a sparsely connected SNN with the same computational efficiency and speed as they can a densely connected SNN because \snm supports both sparse and dense computations.
Lastly, \snm offers support for both single (32-bit) and double (64-bit) precision computations.This gives users the flexibility to achieve higher precision in their computation using double precision or to have more memory at their disposal and thus simulate larger SNNs using single precision.

Lastly, we provide \snm as a free, open-source library on GitHub to foster a collaborative research environment and software development. 
We envision \snm being used by professors, scientists, postdocs, graduate students, undergraduate students, high school students, and SNN enthusiasts all over the world.
We actively encourage contributions from the global SNN and neuromorphic computing community in further developing \snm.

\subsection{Neuron Parameters}

\snm simulates LIF neurons.
Each neuron has an associated index $i \in \mathbb{N}$, and the total number of neurons in the SNN is $N \in \mathbb{N}$.
On desktop and laptop computers, \snm can support up to $N = 10,000$ neurons with a dense implementation and around $N = 100,000$ neurons with a sparse implementation.
We denote the scalar versions of the neuron parameters for neuron $i$ using Greek letters with a subscript $i$ and the vector versions using bold Greek letters.
We denote the time step in square brackets.
For example, the membrane potential of neuron $i$ at time $t$ is denoted as $v_i[t]$.
The membrane potential of all $N$ neurons at time $t$ is denoted as $\mathbf{v}[t]$.
The membrane potential of a LIF neuron is the charge across the membrane of the neuron.
In \snm, the membrane potential represents the internal state of the neuron, and new charges are deposited at every time step.
LIF neurons in \snm have four parameters, as follows:
\begin{enumerate}
    \item \textbf{Threshold} ($\boldsymbol{\nu}$): The threshold is a critical value associated with a neuron:  if the membrane potential of the neuron becomes greater than the threshold, then the neuron can spike. The threshold of neuron $i$ is denoted by $\nu_i$ ($\nu_i \in \mathbb{R}$ $\forall i$), and the threshold of all neurons is denoted by the vector $\boldsymbol{\nu} = [\nu_1 \ \nu_2 \ \ldots \ \nu_N]^T$. The threshold is set when a neuron is created in \snm and does not change during the course of the simulation.
    
    \item \textbf{Leak} ($\boldsymbol{\lambda}$): The leak is the constant amount by which the membrane potential of a neuron is either increased or decreased in order to bring it closer to the reset state of the neuron. If the membrane potential of neuron $i$ is greater than its reset state, then the leak, denoted by $\lambda_i$ ($\lambda_i \in \mathbb{R}$ $\forall i$), will be subtracted from its membrane potential. On the other hand, if the membrane potential of the neuron is less than its reset state, then the leak will be added to its membrane potential. The leak is added to or subtracted from the membrane potential until the membrane potential equals the reset state and no further. The leak of all $N$ neurons in \snm is denoted by the vector $\boldsymbol{\lambda} = [\lambda_1 \ \lambda_2 \ \ldots \lambda_N]^T$ ($\boldsymbol{\lambda} \in \mathbb{R}^N$). The leak is set when a neuron is created and does not change during the course of a simulation.
    
    \item \textbf{Reset State} ($\boldsymbol{\rho}$): The reset state of a neuron is the value that its membrane potential attains after the neuron spikes. It is also the value toward which the membrane potential of the neuron is brought by adding or subtracting leak in case the neuron does not spike. The reset state of neuron $i$ is denoted by $\rho_i$ ($\rho_i \in \mathbb{R}$ $\forall i$), and the reset state of all $N$ neurons is denoted by the vector $\boldsymbol{\rho} = [\rho_1 \ \rho_2 \ \ldots \ \rho_N]^T$ ($\boldsymbol{\rho} \in \mathbb{R}^N$).
    
    \item \textbf{Refractory Period} ($\boldsymbol{\tau}$): The refractory period of a neuron is the number of time steps during which the neuron cannot spike after it spikes at a certain time step. For instance, if a neuron $i$ spikes at time step $t$ and its refractory period is $\tau_i$ ($\tau_i \in \mathbb{W}$ $\forall i$), then the neuron cannot spike for the next $\tau_i$ time steps. The vector of all refractory periods is denoted by $\boldsymbol{\tau} = [\tau_1 \ \tau_2 \ \ldots \ \tau_N]^T$ ($\boldsymbol{\tau} \in \mathbb{W}^N$). To track the refractory period remaining during the course of the simulation, we define a sister variable $\boldsymbol{\tau'} = [\tau'_1 \ \tau'_2 \ \ldots \ \tau'_N]$ such that $\tau'_i \in \mathbb{W}$ $\forall i$ and $\boldsymbol{\tau'} \in \mathbb{W}^N$. Whereas $\boldsymbol{\tau}$ does not change during the course of the simulation, $\boldsymbol{\tau'}$ can change. In fact, $\boldsymbol{\tau'}$ is checked and updated at every time step. A neuron $i$ can spike at time step $t$ only if its refractory period remaining equals $0$ (i.e., $\tau'_i[t] = 0$). 
\end{enumerate}


In \snm, we store each neuron parameter as a Python list. 
When a neuron is created, we append all the neuron parameters to their corresponding lists.
For instance, if a new neuron is created with threshold $\nu$, leak $\lambda$, reset state $\rho$, and refractory period $\tau$, then the lists of thresholds ($\boldsymbol{\nu}$), leaks ($\boldsymbol{\lambda}$), reset states ($\boldsymbol{\rho}$), and refractory periods ($\boldsymbol{\tau}$) would be appended with $\nu$, $\lambda$, $\rho$, and $\tau$, respectively.

\subsection{Synapse Parameters}

\snm does not impose any restrictions on the connectivity or the fan-in/fan-out of the neurons. 
It can simulate all-to-all connectivity in an SNN.
For an SNN with $N$ neurons, the maximum number of synapses is $N^2$, and \snm indeed supports this.
For computational efficiency, we provide two ways of representing the synapses in memory: dense and sparse representations.
The user can select the representation that is the most apt for their application.
\snm also has the capability of automatically selecting dense or sparse representation based on the connectivity of the SNN.
Each synapse has a unique index $(i,j)$, where $i$ and $j$ are the neuron indices of the pre-synaptic and post-synaptic neurons, respectively.
We support three synapse parameters as follows:
\begin{enumerate}
    \item \textbf{Weight} ($\mathbf{W}$): The synaptic weight represents the strength of the connection between two neurons. Each synapse multiplies the incoming spike by its weight. The weight of synapse $(i,j)$ is denoted by $\omega_{i,j}$ ($\omega_{i,j} \in \mathbb{R}$). The matrix of all synaptic weights is denoted by $W$ ($W \in \mathbb{R}^{N \times N}$).
    
    \item \textbf{Delay} ($\mathbf{D}$): The synaptic delay represents the number of time steps it takes for a spike to travel from a pre-synaptic neuron to a post-synaptic neuron. Each synapse has a default delay of $1$, as we assume that it takes at least one unit of time for a synapse to propagate the spike. This assumption is fundamental to determine the computational complexity of neuromorphic algorithms \cite{date2021computational}. The delay of synapse $(i,j)$ is denoted by $\delta_{i,j}$ ($\delta_{i,j} \in \mathbb{N}$). The matrix containing all synaptic delays is denoted by $D$ ($D \in \mathbb{N}^{N \times N}$).
    
    \item \textbf{STDP Enabled Flag} ($\mathbf{E}$): We allow the option of enabling or disabling STDP learning on each synapse in \snm. This is captured by the STDP enabled flag, which for synapse $(i,j)$ is denoted by $\epsilon_{i,j}$ ($\epsilon_{i,j} \in \mathbb{B}$). STDP is enabled if the flag is set to $1$; otherwise, it is disabled. The matrix of STDP enabled flags for all synapses is denoted by $E$ ($E \in \mathbb{B}^{N \times N}$).
\end{enumerate}

While creating synapses, we store the synapse parameters in Python lists. 
When a synapse is created, we append the synapse parameters of the newly created synapse to their corresponding lists.
Let us say a new synapse is to be created with weight $\omega$, delay $\delta$, and STDP enabled flag $\epsilon$. 
If the delay is $1$, then all synapse parameters ($\omega$, $\delta$, and $\epsilon$) get appended to their respective lists: that is, weights list ($\boldsymbol{\omega}$), delays list ($\boldsymbol{\delta}$), and STDP enabled flags list ($\boldsymbol{\epsilon}$).
If the delay $\delta$ is greater than $1$, then \snm creates a chain of additional $\delta - 1$ neurons such that all synapses along this chain have a delay of $1$. 
The last synapse in this chain is assigned the weight $\omega$ and STDP enabled flag $\epsilon$.
All synapse parameters are updated in their corresponding lists during the creation of these chained synapses.
We also store the pre-synaptic and post-synaptic neuron indices for all synapses while creating them.
While running the simulation, synapses are represented using dense or sparse arrays.
In the dense representation, the synapse parameters are stored as $\mathcal{O}(N^2)$ numpy arrays.
In the sparse representation, the synapse parameters are stored as $\mathcal{O}(S)$ sparse scipy arrays.

\subsection{Input Spikes}

In \snm, we enable the user to provide external input spikes to neurons at any time step.
Unlike the `internal' binary spikes propagated by the neurons in the SNN, the `external' spikes are provided by the user and can be real valued. 
We denote the external spike given to neuron $i$ at time step $t$ by $x_i[t]$.
A vector of all input spikes at time step $t$ is denoted by $\mathbf{x}[t]$.
These input spikes are ordered by the time step. 
For the most efficient simulation, it is better to provide details about input spikes before the simulate function is called.

\subsection{Learning Parameters}

\snm supports STDP learning with the following three parameters.
\begin{enumerate}
    \item \textbf{STDP Time Steps} ($T_S$): This is the number of time steps over which STDP operations occur. Typically, $T_S \in \mathbb{W}$. If $T_S$ equals $0$, then no STDP operations occur.
    
    \item \textbf{Coefficients for Positive Updates} ($\boldsymbol{\alpha^+}$): This is a $T_S$ dimensional vector containing coefficients for positive weight updates. Typically, $\alpha^+_i \in \mathbb{R}$ $\forall i\in \{1,2,\ldots,T_S\}$, $\boldsymbol{\alpha^+} \in \mathbb{R}^{T_S}$, and $\alpha_1^+ \ge \alpha_2^+ \ge \ldots \ge \alpha_{T_S}^+ \ge 0$.
    
    \item \textbf{Coefficients for Negative Updates} ($\boldsymbol{\alpha^-}$): $\boldsymbol{\alpha}^-$ This is a $T_S$ dimensional vector of coefficients for negative weight updates in the STDP operation. Typically, $\alpha_i^- \in \mathbb{R}$ $\forall i \in \{1, 2, \ldots, T_S\}$, $\boldsymbol{\alpha^-} \in \mathbb{R}^{T_S}$, and $\alpha_1^- \le \alpha_2^- \le \ldots \le \alpha_{T_S}^- \le 0$.
\end{enumerate}

If the user only requires positive weight updates during STDP, then $\alpha^-_i$ can be set to $0$ for all $i$.
Similarly, $\alpha^+_i$ can be set to $0$ for all $i$ if only negative weight updates are required during STDP.

\subsection{Modeling the LIF Operations}
\label{sub:modeling}

We model the LIF operations using a matrix-based approach. To the best of our knowledge, this approach is unprecedented in the literature.
We use a matrix-based approach in order cater to the processors available on today's laptops and desktops.
The CPUs available on these machines are highly optimized to run matrix or vector operations using single instruction, multiple data (SIMD) computations.
By modeling the LIF operations using matrices and vectors, and by leveraging the numpy and scipy libraries, which are based on the BLAS library in C \cite{lawson1979basic}, we were able to perform the LIF operations in a highly optimized and computationally efficient manner.

In this section, we derive the discretized version of Equation \ref{eq:lif} to make it amenable to computation on CPUs.
Firstly, we start by multiplying Equation \ref{eq:lif} by $\frac{dt}{\tau_m}$ on both sides.
\begin{align}
    dV(t) = \left[ \frac{-(V(t) - V_0) + R I(t)}{\tau_m} \right] dt
\end{align}

Next, we discretize the equation in time and use square brackets to make this evident.
Note that we switch to using square brackets for discrete time $t$ according to our notation described in Section \ref{sec:notation}.
\begin{align}
    \Delta V[t] = \left[ \frac{-(V[t - \Delta t] - V_0) + R I[t - \Delta t]}{\tau_m} \right] \Delta t
\end{align}

Next, we choose a small enough time step such that $\Delta t$ equals $1$. In real physical units, this time difference can vary from application to application. Some applications may have a time step of 1~ms whereas others might have a time step of 1~ns or 1~ps.
\begin{align}
    \Delta V[t] = \frac{-(V[t-1] - V_0) + R I[t-1]}{\tau_m}
\end{align}

Now, we move the $\tau_m$ into the two terms on the right-hand side.
\begin{align}
    \Delta V[t] = \underbrace{\frac{-(V[t-1] - V_0)}{\tau_m}}_{\text{leak}} + \underbrace{\frac{R}{\tau_m}}_{\text{weights}} \underbrace{I[t-1]}_{\text{spikes}}
\end{align}

This separates the leak term from the weights and spikes term. 
We further split the second term on the right-hand side into two terms, one for the input spikes sent by the user and the other for the spikes propagated within the network.
\begin{align}
    \Delta V[t] = \underbrace{\frac{-(V[t-1] - V_0)}{\tau_m}}_{\text{leak}} + \underbrace{\frac{R I_{in}[t-1]}{\tau_m}}_{\text{input spikes}} + \underbrace{\frac{R I_{net}[t-1]}{\tau_m}}_{\text{network spikes}}
\end{align}

Now, let us rewrite the above equation using the \snm notation from Table \ref{tab:neuron-notation} and Table \ref{tab:synapse-notation}. 
The membrane potential of neuron $i$ at time step $t$ in \snm is denoted using $v_i[t]$.
Therefore, the $\Delta V[t]$ term becomes $v_i[t] - v_i[t-1]$.
\snm supports a constant leak at each time step.
Therefore, the leak term for neuron $i$ simply reduces to $-\lambda_i$.
The input spikes in \snm are denoted by $x_i[t]$, which are not binary spikes but, rather, can have a real number value.
Therefore, the input spikes term becomes $x_i[t]$.
Lastly, the network spikes term can be split into $\frac{R}{\tau_m}$, which denotes the synaptic weights of all the incoming synapses and the $I_{net}[t]$, which denotes the binary spikes along all incoming synapses.
Therefore, the above equation can be rewritten using the \snm notation as follows:
\begin{align}
    v_i[t] \ - \ v_i[t-1] \ = \ -\lambda_i \ + \ x_i[t] \ + \ \sum_{j=1}^{N} \omega_{i,j}[t-1] \ s_j[t-1]
\end{align}

Note that the summation of $j$ is taken over all the neurons in the SNN.
If a synapse from neuron $j$ to neuron $i$ does not exist in the SNN, then we set its weight to $0$.
From a computational efficiency point of view, it is more efficient to perform this operation over weights that are initialized to $0$ rather than searching for only those synapses that exist in the SNN.
We now move the $v_i[t-1]$ term to the right-hand side to yield the equation to update the membrane potential of neuron $i$ at time step $t$:
\begin{align}
    v_i[t] \ = \ v_i[t-1] \ - \ \lambda_i \ + \ x_i[t] \ + \ \sum_{j=1}^{N} \omega_{i,j}[t-1] \ s_j[t-1]
\end{align}

We now vectorize this equation to compute the membrane potential of all $N$ neurons in one shot:
\begin{align}
    \mathbf{v}[t] \ = \ \mathbf{v}[t-1] \ - \ \boldsymbol{\lambda} \ + \ \mathbf{x}[t] \ + \ W^T[t-1] \ \mathbf{s}[t-1]  \label{eq:membrane-potential-update}
\end{align}

Note that in the matrix $W$, each element $w_{i,j}$ denotes the synaptic weight of the synapse going \emph{from} neuron $i$ \emph{to} neuron $j$.
In order to match the weights to the incoming spike vector $s$, we must take the transpose of $W$.
Equation \ref{eq:membrane-potential-update} gives us a vectorized equation to update the membrane potential of all neurons at time step $t$. 

We would like to point out a few considerations about computing the leak in Equation \ref{eq:membrane-potential-update}.
The leak tries to bring the membrane potential of the neuron back to the rest state.
If the membrane potential $v_i[t-1]$ is greater than the reset states $\rho_i$ for neuron $i$, then the leak is subtracted as shown in Equation \ref{eq:lif}.
However, if $v_i[t-1] < \rho_i$ for some neuron $i$, then the constant leak $\lambda_i$ will be \emph{added} to the membrane potential instead of being \emph{subtracted} in order to bring the membrane potential closer to the reset state.
Another point to keep in mind about the leak is that after subtracting/adding leak to the membrane potential, if the membrane potential becomes lower/higher than the reset state, then we set the membrane potential to the reset state.

The neuron spikes if its membrane potential is greater than its threshold and its refractory period remaining at the current time step is $0$.
If the neuron spikes, then it enters its refractory period. 
We do this by setting its refractory period remaining $\tau_i'$ to its refractory period $\tau_i$.
If the neuron does not spike, we decrement its refractory period by $1$ until it reaches $0$.
Accordingly, we have the two following equations to compute the spikes and the refractory periods.
\begin{align}
    s_i[t]  &= 
        \begin{cases}
            1 &\qquad \text{if} \ \ v_i[t] > \nu_i \ \ \text{and} \ \ \tau_i'[t] = 0 \\
            0 &\qquad \text{otherwise}
        \end{cases}    \label{eq:spike} \\
    \tau_i'[t] &= 
        \begin{cases}
            \tau_i                              & \ \text{if} \ \ s_i[t] = 1  \\
            \max \{ 0, \ \tau_i'[t-1] - 1 \}    & \ \text{otherwise}
        \end{cases}    \label{eq:refractory}
\end{align}
where the $\max\{...\}$ function selects the maximum element out of all the elements in a given set.

\subsection{Modeling the STDP Operations}

The STDP algorithm is often colloquially summarized as follows: \emph{neurons that fire together, wire together}.
More formally, the STDP algorithm can be described as follows. 
Let us say that a neuron $i$ spiked at time step $t-1$ and another neuron $j$ spiked at time step $t$ (i.e., neuron $j$ spiked immediately \emph{after} neuron $i$).
In this case, the weight of synapse $(i,j)$ should be increased.
The rationale behind this is that the spiking behavior of neuron $i$ is positively correlated with that of neuron $j$.
In some cases, neuron $i$ could even be thought of as \emph{causing} neuron $j$ to spike, although causality is not always guaranteed.
Conversely, if neuron $j$ spiked at time step $t-1$ and neuron $i$ spiked at time step $t$---that is, if neuron $j$ spiked before neuron $i$---then the weight of synapse $(i,j)$ will be decreased.
This is because neuron $j$ spiked before neuron $i$, meaning their spiking behavior is negatively correlated.

Now, in a series of steps, we will formulate the STDP weight update as matrix--vector operations.
This is how we implement STDP operations in \snm.
First, we want to find the indices of the weights that need to be updated.
To do so, we need to look at the spike vectors at times $t-1$ and $t$.
Since the spike vectors are binary ($s[t-1]$, $s[t] \in \mathbb{B}^N$), their outer product gives us an $N \times N$ binary matrix.
The non-zero entries of this matrix represent the weights for which the pre-synaptic neuron spiked at time step $t-1$ and the post-synaptic neuron spiked at time step $t$.
This is a positive correlation, as described above, and these weights must be increased.
Thus, the positive weight update matrix at time step $t$, $\Delta W^+ [t]$ can be written as follows.
\begin{align}
    \Delta W^+ [t] = \mathbf{s}[t-1] \ \mathbf{s}^T[t] 
\end{align}

Note that if we subtract $\Delta W^+ [t]$ from an $N \times N$ matrix of ones, we will get the indices of the weights that should be decreased.
Thus, the negative weight update matrix at time step $t$, $\Delta W^- [t]$ can be written as follows.
\begin{align}
    \Delta W^- [t] = 1_{N \times N} - \mathbf{s}[t-1] \ \mathbf{s}^T[t] 
\end{align}
Typically, $\Delta W^+$ and $\Delta W^-$ are multiplied by coefficients $\alpha_1^+$ and $\alpha_1^-$, respectively.
These coefficients are called \textit{potentiation amplitude} and \textit{depression amplitude}, respectively, in neuroscience.
They are described in Table \ref{tab:stdp-notation}.
This is akin to the learning rate used in the backpropagation algorithm for training ANNs. 
Accounting for these coefficients, we can rewrite $\Delta W^+[t]$ and $\Delta W^-[t]$ as follows.
\begin{align}
    \Delta W^+ [t] &= \alpha_1^+ \ \mathbf{s}[t-1] \ \mathbf{s}^T[t] \\
    \Delta W^- [t] &= \alpha_1^- \ \left(1_{N \times N} - \mathbf{s}[t-1] \ \mathbf{s}^T[t] \right)
\end{align}

Furthermore, STDP operations occur over $T_S$ time steps.
Accordingly, we have $T_S$ amplitudes of potentiation and depression, $\alpha_1^+, \alpha_2^+, \ldots, \alpha_{T_S}^+$ and $\alpha_1^-, \alpha_2^-, \ldots, \alpha_{T_S}^-$.
These are applied to the outer products of the spike vectors over previous $T_S$ time steps. 
The resulting matrices are summed together to yield an updated expression for the positive and negative weight updates as follows.
\begin{align}
    \Delta W^+ [t] &= \sum_{t_s=1}^{T_S} \alpha_{t_s}^+ \ \mathbf{s}[t- t_s] \ \mathbf{s}^T[t] \label{eq:positive-update} \\
    \Delta W^- [t] &= \sum_{t_s=1}^{T_S} \alpha_{t_s}^- \ \left(1_{N \times N} - \mathbf{s}[t-t_s] \ \mathbf{s}^T[t] \right) \label{eq:negative-update}
\end{align}

Essentially, for positive weight updates, we apply the first potentiation amplitude $\alpha_1^+$ to the most recent spike vector, $\mathbf{s}[t-1]$, the second potentiation amplitude $\alpha_2^+$ to the second most recent spike vector, $\mathbf{s}[t-2]$, and so on.
Typically, we have $\alpha_1^+ \ge \alpha_2^+ \ge \ldots \ge \alpha_{T_S}^+ \ge 0$, giving more importance to the recent spike vectors and less importance to older spike vectors.
We eventually add all these matrices, and that becomes our positive weight update matrix at time step $t$.
The same can be said for the negative weight updates, where we have $\alpha_1^- \le \alpha_2^- \le \ldots \le \alpha_{T_S}^- \le 0$.

Given $W^+[t]$ and $W^-[t]$, we can write the overall weight update $\Delta W$ as follows.
Note that we do not need to subtract $\Delta W^-$ because the depression coefficients are already assumed to be negative. 
\begin{align}
    \Delta W[t] = \Delta W^+[t] + \Delta W^-[t]
\end{align}

Next, \snm allows users to enable STDP on certain synapses and disable it on others.
We have a variable $\epsilon_{i,j}$ associated with synapse $(i,j)$ to capture this information.
If STDP is enabled on synapse $(i,j)$, then $\epsilon_{i,j}$ is set to $1$, else $0$.
This information is also represented in the matrix $E$, where the element in the $i^{\text{th}}$ row and $j^{\text{th}}$ column is $\epsilon_{i,j}$.
$E$ is called the STDP enabled matrix, and $E \in \mathbb{B}^{N \times N}$.
To allow STDP operations on only those synapses for which STDP is enabled, we need to multiply the weight updates by $E$ element-wise.
We denote the element-wise multiplication operator for two matrices using the $\odot$ symbol.
The resulting matrix is added to the weight matrix $W$ to update the weights at time step $t$.
Therefore, the overall weight update can be written as follows.
\begin{align}
    W[t] = W[t-1] + E \odot \Delta W[t]     \label{eq:stdp-weight-update}
\end{align}

\subsection{Setting Up the Simulation Environment}

All the neuron parameters, synapse parameters, and learning parameters are stored in memory as Python lists when the SNN is being created. 
The input spikes are stored as a Python dictionary, keyed on the time steps for efficient retrieval during simulation. 
When the simulation environment is being initialized, numpy arrays are created for all neuron and learning parameters.
If the representation of matrices pertaining to the synaptic parameters is dense, then the synaptic weights and STDP enabled flags are stored as numpy arrays.
If the representation is sparse, then they are stored as scipy sparse arrays. 
We also initialize variables to store the spikes and the spike trains.

\subsection{Simulation Algorithm}

The simulation algorithm implements Equation \ref{eq:membrane-potential-update}, Equation \ref{eq:spike}, Equation \ref{eq:refractory}, and Equation \ref{eq:stdp-weight-update} at each time step.
We leverage numpy and scipy arrays to store the vectors and matrices in memory.
We also implement the computations in a manner that is amenable for SIMD computations on hardware. 
We assume here that the simulation is run for $T$ time steps and determine the computational complexity of updating the membrane potential of the neurons (Equation \ref{eq:membrane-potential-update}) and updating the synaptic weights using STDP (Equation \ref{eq:stdp-weight-update}).
These two equations are the dominant computations that must be performed at each simulation time step.

For storing the vectors in Equation \ref{eq:membrane-potential-update}---$\mathbf{v}$, $\boldsymbol{\lambda}$, $\mathbf{x}$, $\mathbf{s}$ etc.---we need $\mathcal{O}(N)$ space.
For storing the matrix $W$ under a dense implementation, we need $\mathcal{O}(N^2)$ space because with $N$ neurons, the maximum number of synapses possible is $N^2$.
Under a sparse implementation, storing $W$ would require $\mathcal{O}(S)$ space.
Therefore, the space complexity in a dense implementation is $\mathcal{O}(N^2)$, and that in a sparse implementation is $\mathcal{O}(N + S)$.
To compute the vector operations, we would need $\mathcal{O}(N)$ time.
To compute the matrix operations under a dense implementation, we would need $\mathcal{O}(N^2)$ time, and under a sparse implementation, we would need $\mathcal{O}(S)$ time.
Therefore, the overall time complexity under a dense implementation is $\mathcal{O}(N^2)$, and that in a sparse implementation is $\mathcal{O}(N + S)$.

The space complexity for computing Equation \ref{eq:stdp-weight-update} is $\mathcal{O}(N^2)$ for a dense implementation and $\mathcal{O}(S)$ for a sparse implementation.
Since all the vectors are stored as dense vectors in \snm, the time complexity of computing the outer products in Equation \ref{eq:positive-update} and Equation \ref{eq:negative-update} is $\mathcal{O}(N^2)$.
We compute $T_S$ such outer products at each time step.
Therefore, the overall time complexity of computing Equation \ref{eq:positive-update} and Equation \ref{eq:negative-update} is $\mathcal{O}(N^2 \cdot T_S)$.
The time complexity of computing Equation \ref{eq:stdp-weight-update} is $\mathcal{O}(N^2)$ for a dense implementation and $\mathcal{O}(S)$ for a sparse implementation.
However, the overall time complexity of STDP computations is dominated by Equation \ref{eq:positive-update} and Equation \ref{eq:negative-update}.
Therefore, the overall time complexity of STDP computations is $\mathcal{O}(N^2 \cdot T_S)$, and the overall space complexity of STDP computations is $\mathcal{O}(N^2)$ for dense and $\mathcal{O}(S)$ for sparse implementations. 

We will also need $\mathcal{O}(N \cdot T)$ space to store the spike train.
Therefore, the overall space complexity for dense computations is $\mathcal{O}(N^2) + \mathcal{O}(N^2 \cdot T_S) + \mathcal{O}(N \cdot T)$.
The overall space complexity for sparse computations is $\mathcal{O}(S) + \mathcal{O}(N^2 \cdot T_S) + \mathcal{O}(N \cdot T)$.
Taking the highest-order terms, we get overall space complexity as $\mathcal{O}(N^2 \cdot T_S + N \cdot T)$.
Similarly, the overall time complexity for each time step is $\mathcal{O}(N^2 \cdot T_S)$ for both sparse and dense computations.
This is dominated by the STDP operations.
The overall time complexity across $T$ simulation time steps is $\mathcal{O}(N^2 \cdot T_S \cdot T)$.
Note that since all operations are modeled using matrices and vectors, they can be parallelized to take advantage of multiple CPUs on a compute platform.
Numpy and scipy inherently support this parallel execution because they are based on the BLAS library in C. 
Although the time complexity is a polynomial expression in the number of inputs  and is thus considered efficient, we see even better performance in practice because of the parallel execution.

\section{Results}
\label{sec:results}

\snm supports both sparse and dense representations of the SNN.
We conducted a detailed study comparing the total run time and memory consumption for both sparse and dense representations of various SNNs generated uniformly at random, as detailed in in Section \ref{sub:comp-efficiency}.
We also compared the performance of \snm with four SNN simulators: NEST, Brian2, BindsNET, and snnTorch.
As shown in Section \ref{sub:benchmarks}, we used \snm in two conventional benchmark ML examples (Digits and Citation Networks), two neuromorphic ML examples (N-CARS and ASL-DVS), and two non-ML examples (shortest path and arithmetic primitives).
These problems are crucial for general-purpose computing applications of neuromorphic computing.

\subsection{Computational Efficiency}
\label{sub:comp-efficiency}

All experiments in this section were conducted on a KVM virtual machine running Rocky Linux 9.8 with 32 vCPUs and 128 GiB of RAM, hosted on an AMD EPYC 7702 64-core (Zen 2) processor. 
Simultaneous multithreading was disabled and a single NUMA node was used. 
Storage consisted of a 60 GiB virtio block device holding the root file system as well as a 1.2 TiB shared NFS v4.2 volume mounted for user data. Transparent huge pages were enabled in always mode, and no swap was configured, so no measurement was affected by paging.

\subsubsection{Performance of Sparse and Dense Simulation Modes}

\begin{figure}[t!]
    \centering
    \begin{subfigure}{0.49\textwidth}
        \centering
        \includegraphics[width=\textwidth]{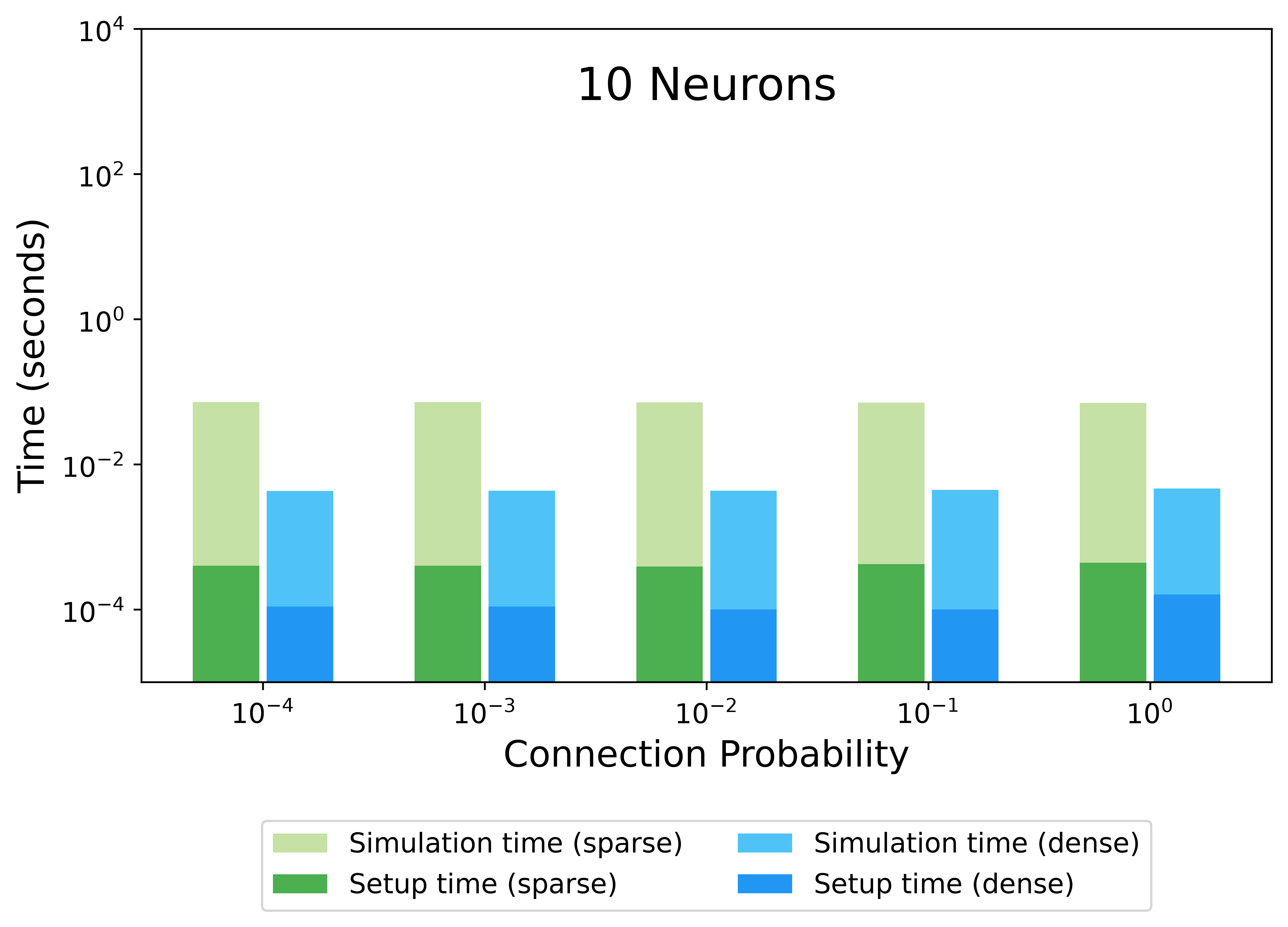}
        \caption{10 Neurons}
        \label{subfig:10-neurons}
    \end{subfigure}
    \begin{subfigure}{0.49\textwidth}
        \centering
        \includegraphics[width=\textwidth]{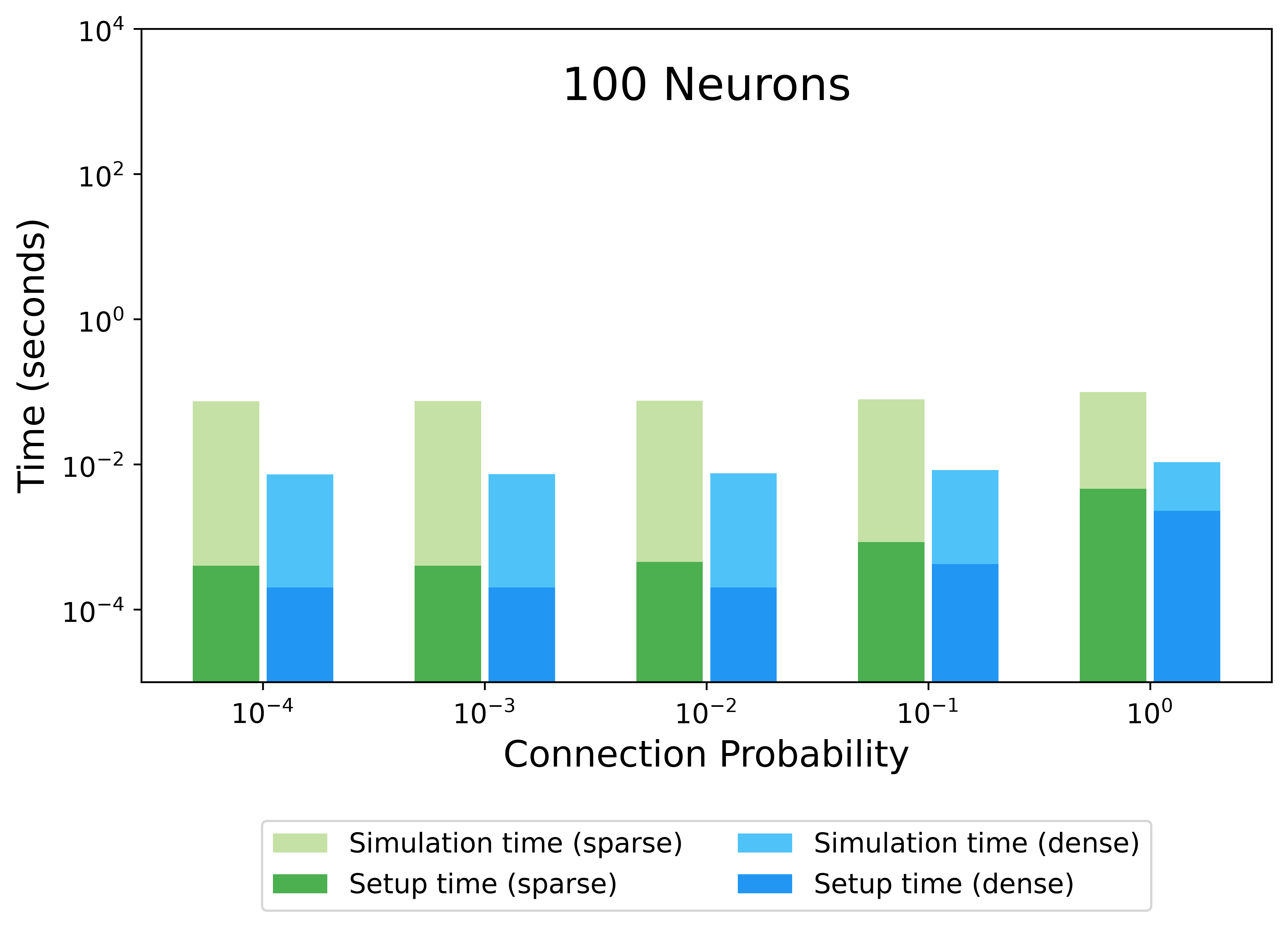}
        \caption{100 Neurons}
        \label{subfig:100-neurons}
    \end{subfigure}
    \\
    \begin{subfigure}{0.49\textwidth}
        \centering
        \includegraphics[width=\textwidth]{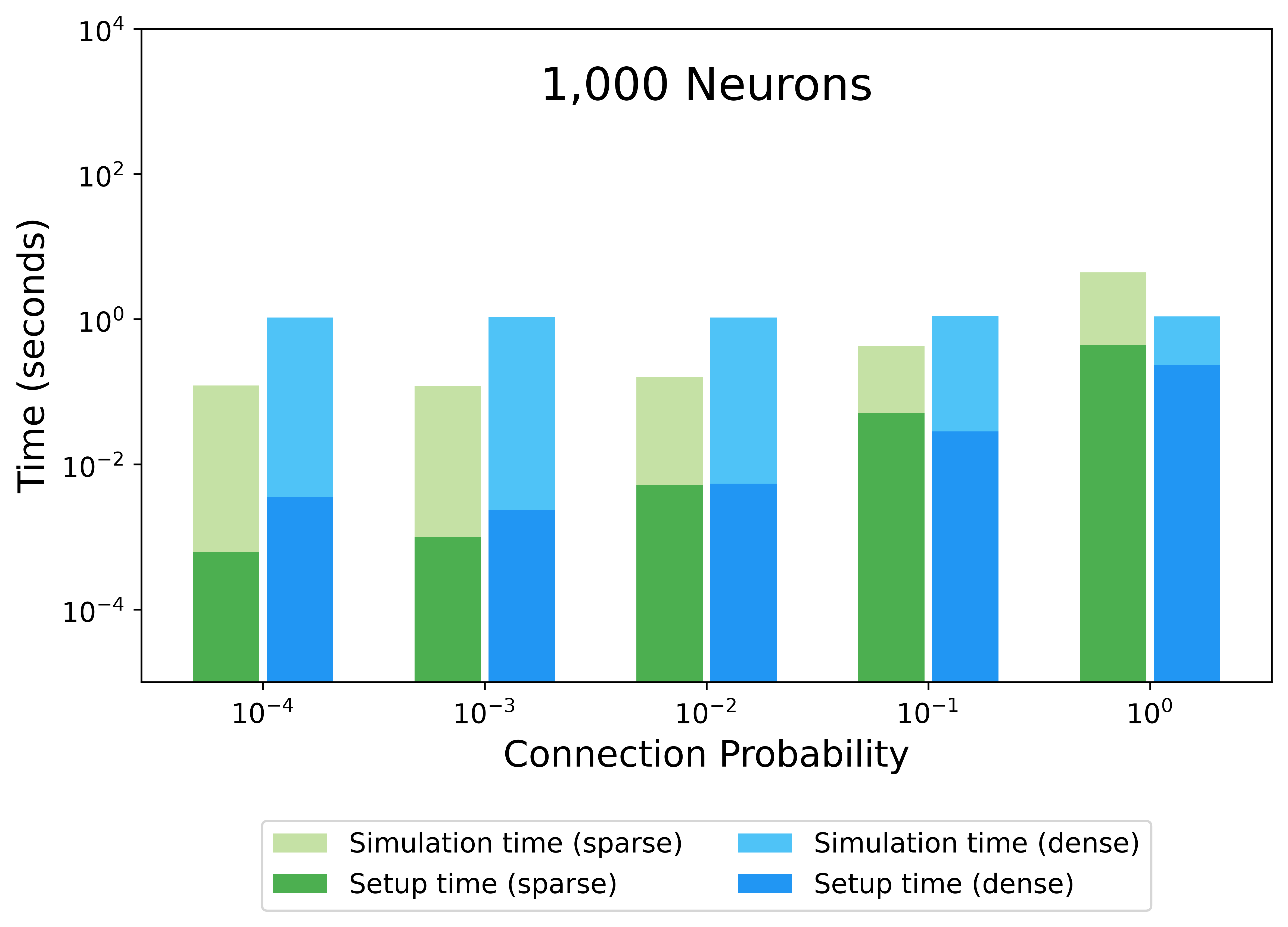}
        \caption{1,000 Neurons}
        \label{subfig:1000-neurons}
    \end{subfigure}
    \begin{subfigure}{0.49\textwidth}
        \centering
        \includegraphics[width=\textwidth]{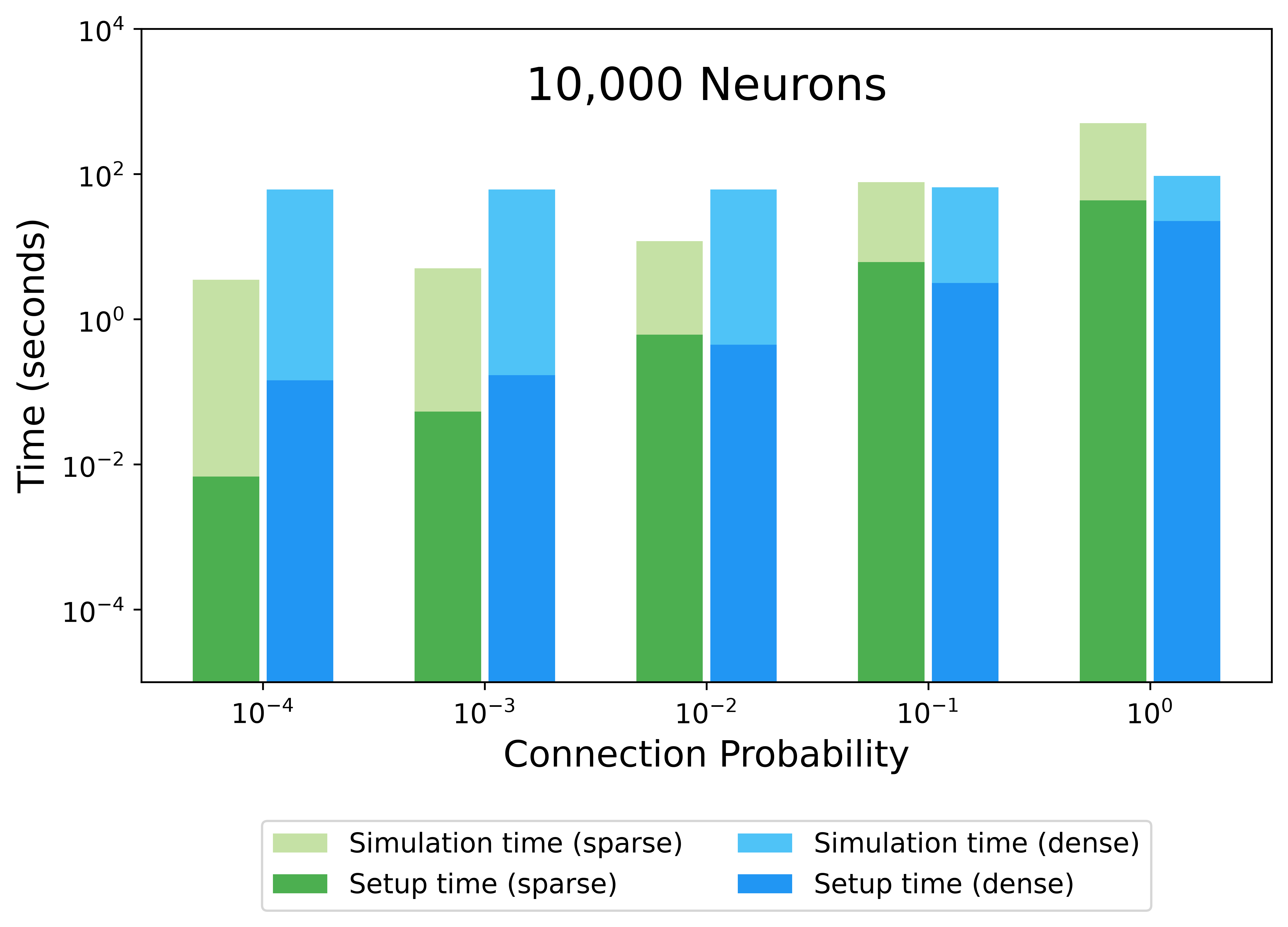}
        \caption{10,000 Neurons}
        \label{subfig:10000-neurons}
    \end{subfigure}
    \caption{Comparison of run times of sparse and dense modes in \snm for (a) 10 neurons; (b) 100 neurons; (c) 1,000 neurons; and (d) 10,000 neurons. The connection probability is shown on the X-axis, and the total run time in seconds is shown on the Y-axis.}
    \label{fig:svd-runtimes}
\end{figure}

\begin{figure}[t!]
    \centering
    \begin{subfigure}{0.49\textwidth}
        \centering
        \includegraphics[width=\textwidth]{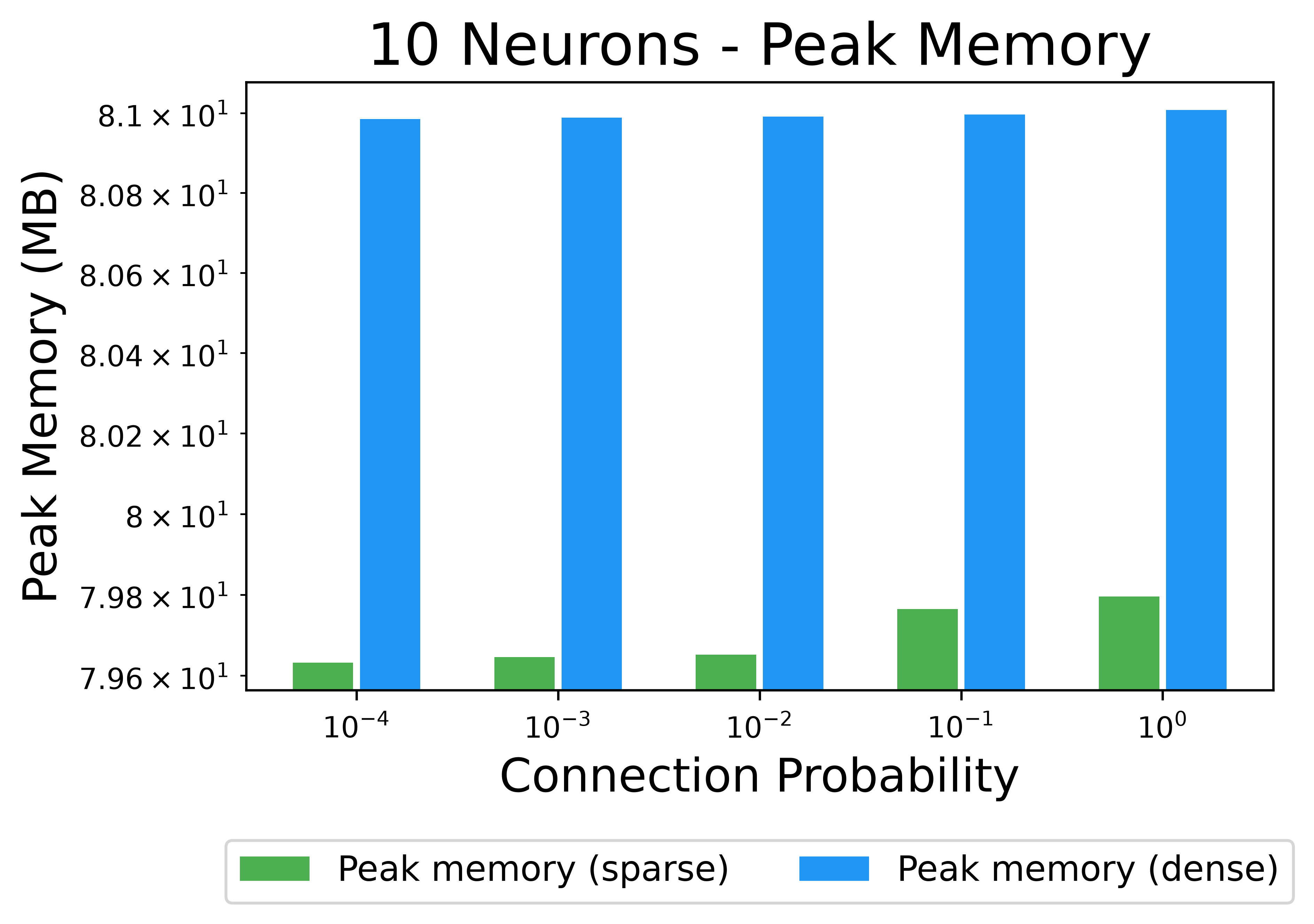}
        \caption{10 Neurons}
        \label{subfig:10-neurons}
    \end{subfigure}
    \begin{subfigure}{0.49\textwidth}
        \centering
        \includegraphics[width=\textwidth]{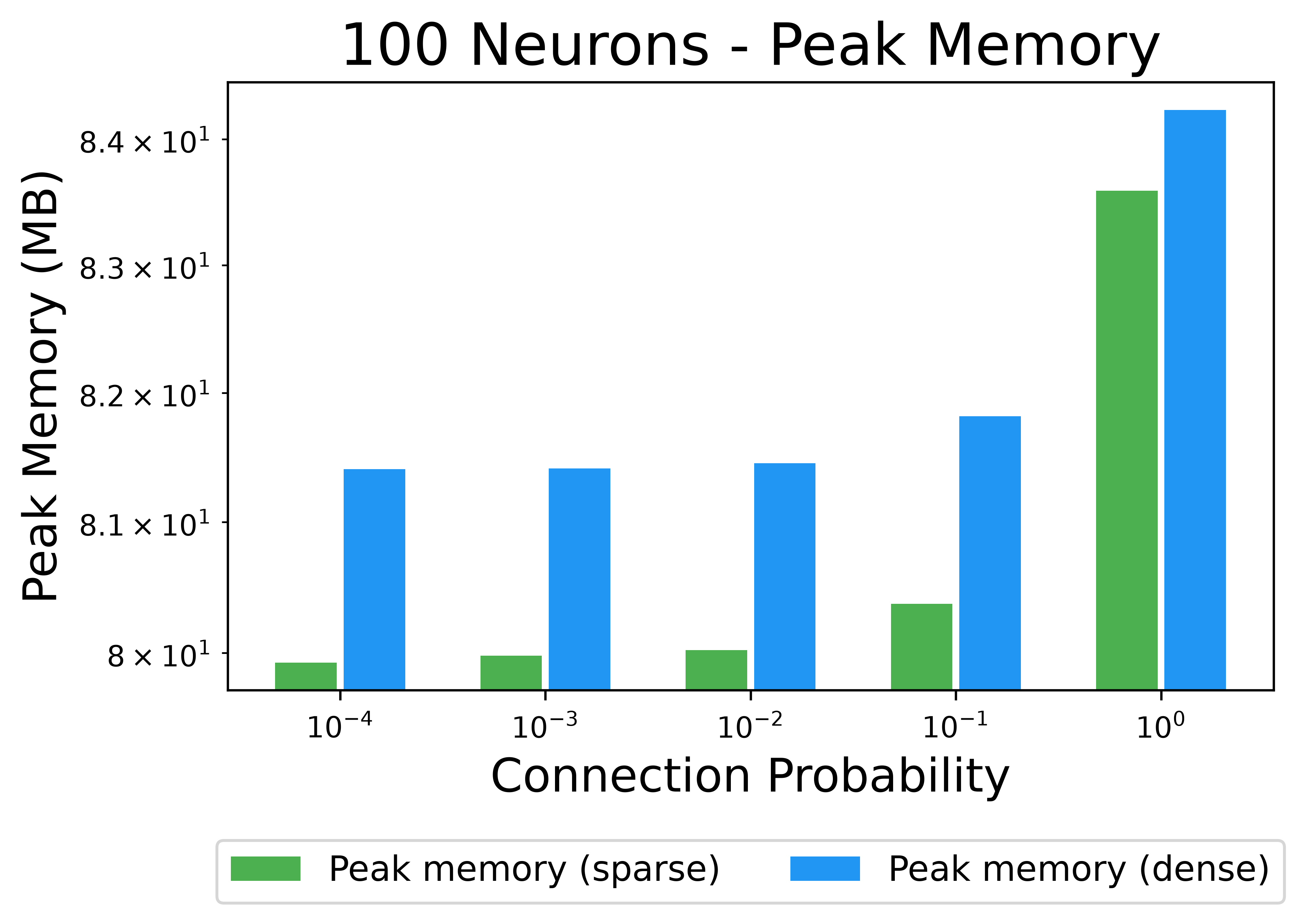}
        \caption{100 Neurons}
        \label{subfig:100-neurons}
    \end{subfigure}
    \\
    \begin{subfigure}{0.49\textwidth}
        \centering
        \includegraphics[width=\textwidth]{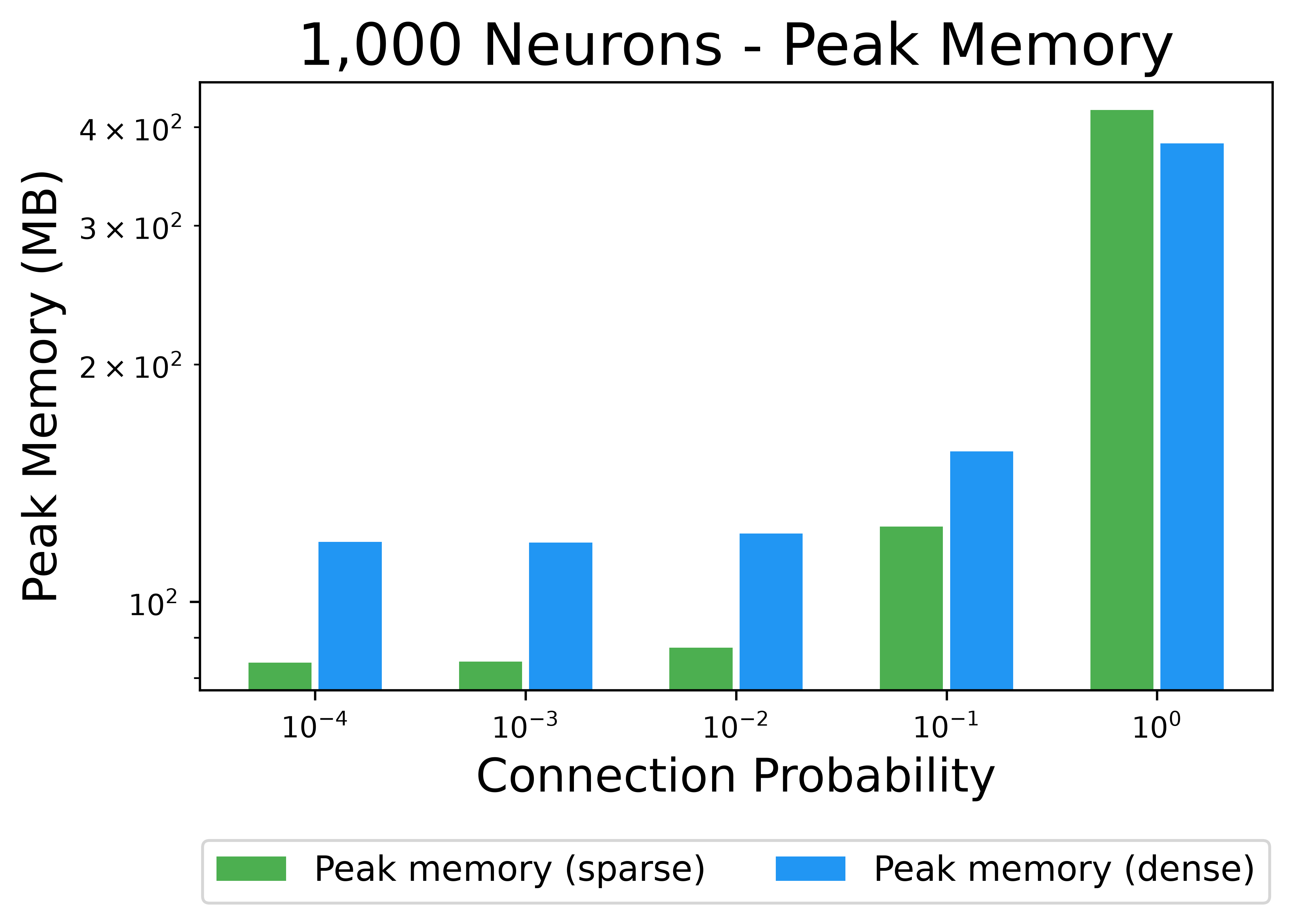}
        \caption{1,000 Neurons}
        \label{subfig:1000-neurons}
    \end{subfigure}
    \begin{subfigure}{0.49\textwidth}
        \centering
        \includegraphics[width=\textwidth]{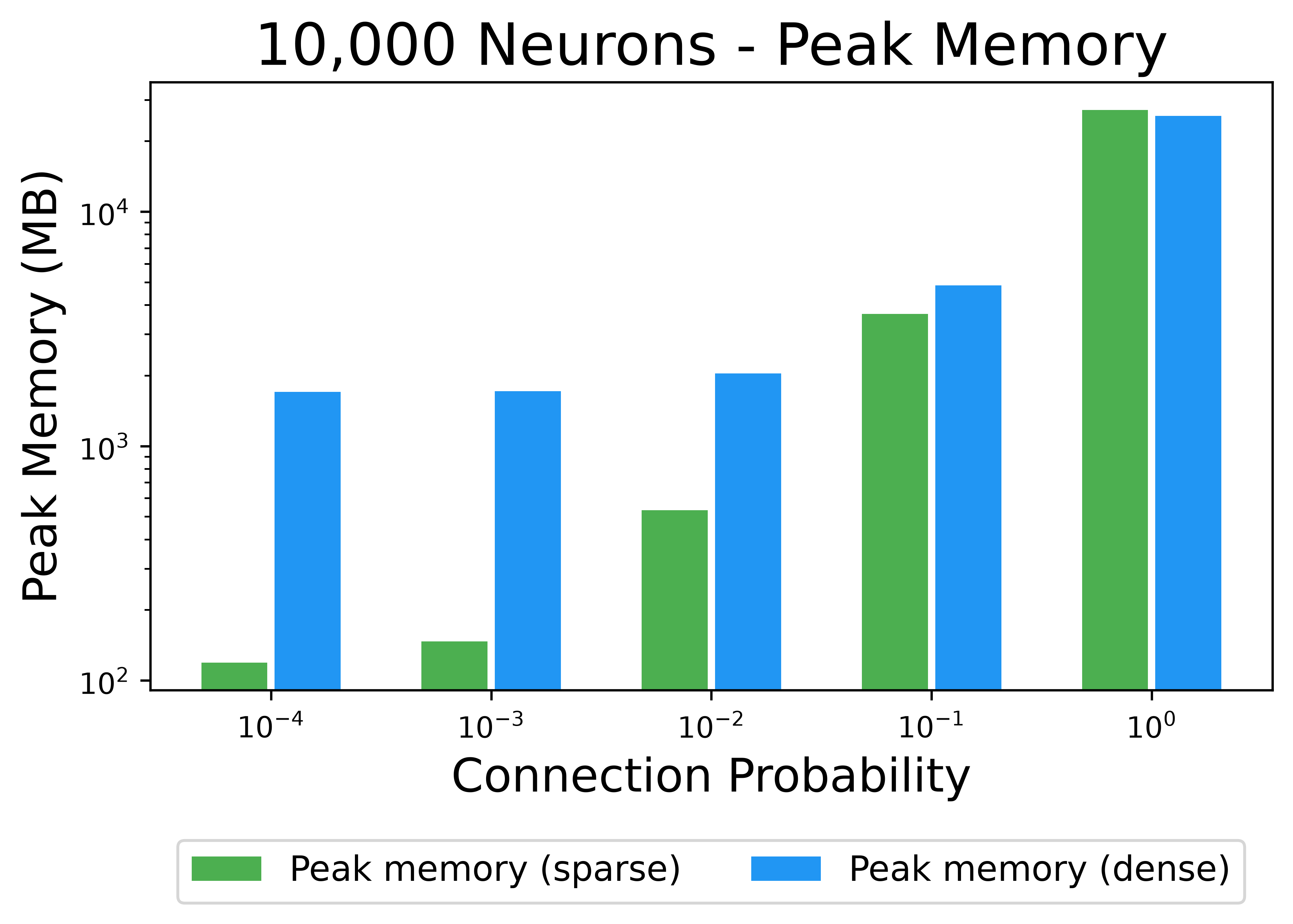}
        \caption{10,000 Neurons}
        \label{subfig:10000-neurons}
    \end{subfigure}
    \caption{Comparison of memory consumed by the sparse and dense modes in \snm for (a) 10 neurons; (b) 100 neurons; (c) 1,000 neurons; and (d) 10,000 neurons. The connection probability is shown on the X-axis, and the total memory consumed in megabytes (MB) is shown on the Y-axis.}
    \label{fig:svd-memory}
\end{figure}

We first compare the performance of the sparse and dense execution modes in \snm.
In this section, the connection probability is defined as $S/N^{2}$, where $S$ is the number of synapses, and $N$ is the number of neurons in the SNN.
Since $N^{2}$ is the maximum possible number of synapses in an SNN of $N$ neurons, this quantity expresses the fraction of connections that are actually present.
We evaluated a structured grid of four network sizes ($N = 10$, $100$, $1{,}000$, and $10{,}000$) and five connection probabilities ($\frac{S}{N^2} = 10^{-4}$, $10^{-3}$, $10^{-2}$, $10^{-1}$, and $1.0$).
Each of these twenty combinations was simulated in both modes, yielding forty distinct simulation configurations, each of which was executed ten times for a total of four hundred runs.
The values shown in Figure \ref{fig:svd-runtimes} and Figure \ref{fig:svd-memory} are averaged over the ten runs of each configuration.

Each SNN was constructed by first creating $N$ neurons and then creating synapses uniformly at random according to the connection probability of the configuration.
STDP was enabled on all synapses, and input spikes were assigned uniformly at random.
Every network was simulated for 100 time steps.
For each run, we recorded the network setup time, the simulation time, the total run time, and the peak resident memory so that computational and memory costs could be compared directly across the two modes.

Figure~\ref{fig:svd-runtimes} presents the total run time (Y-axis) for both sparse and dense modes as a function of connection probability (X-axis) for each network size; because the Y-axis denotes run time, lower values are better.
Figure \ref{fig:svd-memory} presents the peak memory consumption (Y-axis) for both modes as a function of connection probability (X-axis) for each of the four network sizes; once again, lower values are better.
The results reveal a clear crossover the location of which depends jointly on network size and connectivity.
For the smallest networks (10--100 neurons), dense mode was consistently faster at every connection probability, with simulation times of roughly $0.004$--$0.011$~s compared with $0.07$--$0.11$~s in sparse mode.
Peak memory in this regime was essentially identical between the two modes (approximately 80~MB) and was dominated by run time overhead rather than by the network itself.
At 1{,}000 neurons the ordering reversed at low and moderate connectivity: mean total run time in sparse mode was $5.1\times$ faster at $10^{-4}$ ($0.166$~s vs. $0.851$~s), $3.6\times$ faster at $10^{-3}$ ($0.226$~s vs. $0.821$~s), and $2.4\times$ faster at $10^{-2}$ ($0.400$~s vs. $0.949$~s), with correspondingly lower peak memory (84--88~MB vs. 119--122~MB).
At a connection probability of $10^{-1}$, the two modes were effectively tied in terms of run time ($1.72$~s vs. $1.75$~s), although the sparse mode retained a memory advantage (about 124~MB vs 155~MB), and at full connectivity the dense mode was again faster ($3.23$~s vs $9.82$~s).

The sparse mode performance was the most prominent at the largest scale.
For 10{,}000 neurons, sparse mode reduced total run time by approximately $45\times$ at a connection probability of $10^{-4}$ ($3.93$~s vs. $179$~s), $26\times$ at $10^{-3}$ ($6.67$~s vs. $176$~s), and $8\times$ at $10^{-2}$ ($23.0$~s vs. $184$~s), and it remained faster at $10^{-1}$ ($200$~s vs. $273$~s).
Peak memory usage in these regimes was between $1.3\times$ and $14\times$ lower than in dense mode, with the largest reductions at the sparsest connectivities.
At full connectivity, however, sparse mode required substantially more time ($2{,}140$~s vs. $508$~s) and slightly more memory ($27.2$~GB vs. $25.6$~GB) than the dense mode.

The results above indicate that neither representation dominates across the parameter space.
Dense execution is preferable when the problem is small because its lower per-operation overhead outweighs any redundant computation, as  well as when the network is close to fully connected because the sparse representation traverses almost as many elements as the dense one while paying additional indexing costs.
Sparse execution is preferable for large, genuinely sparse networks, which is the regime most representative of application-scale SNNs; it yielded speedups of up to $45\times$ and memory reductions of up to $14\times$ in our experiments.
Because users can present \snm with any of these configurations, it supports both execution modes.
When the user does not explicitly select a mode, \snm chooses one automatically from the number of neurons and the number of synapses in the SNN. 
Dense mode is selected for small problems and for large, densely connected problems. 
Sparse mode is selected for large, sparse problems.

\subsubsection{Comparison with Other Simulators}

\begin{figure}[t!]
    \centering
    \begin{subfigure}{0.49\textwidth}
        \centering
        \includegraphics[width=\textwidth]{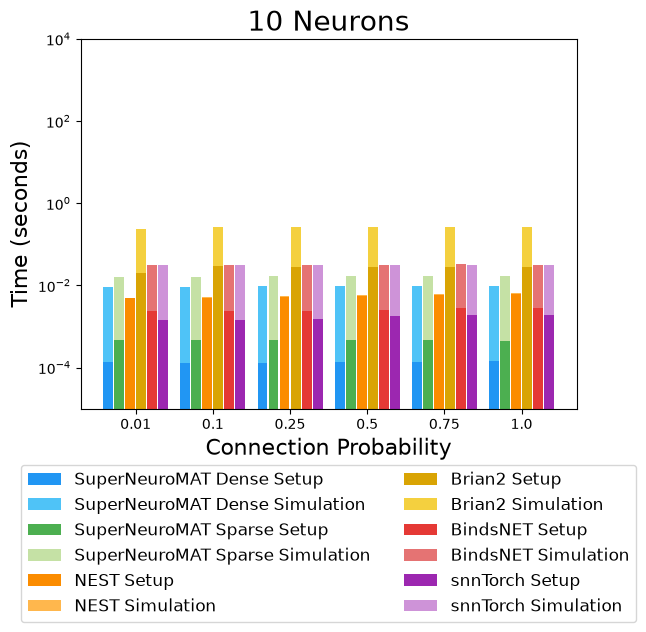}
        \caption{10 Neurons}
        \label{subfig:10-neurons}
    \end{subfigure}
    \begin{subfigure}{0.49\textwidth}
        \centering
        \includegraphics[width=\textwidth]{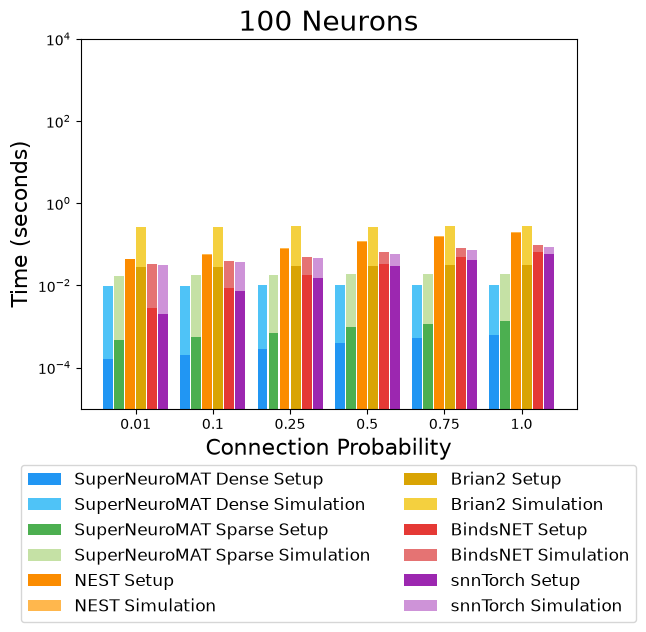}
        \caption{100 Neurons}
        \label{subfig:100-neurons}
    \end{subfigure}
    \\
    \begin{subfigure}{0.49\textwidth}
        \centering
        \includegraphics[width=\textwidth]{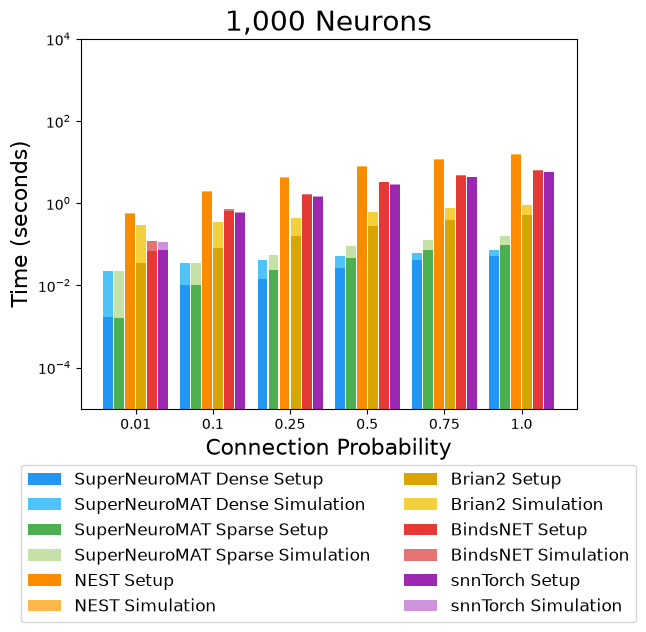}
        \caption{1,000 Neurons}
        \label{subfig:1000-neurons}
    \end{subfigure}
    \begin{subfigure}{0.49\textwidth}
        \centering
        \includegraphics[width=\textwidth]{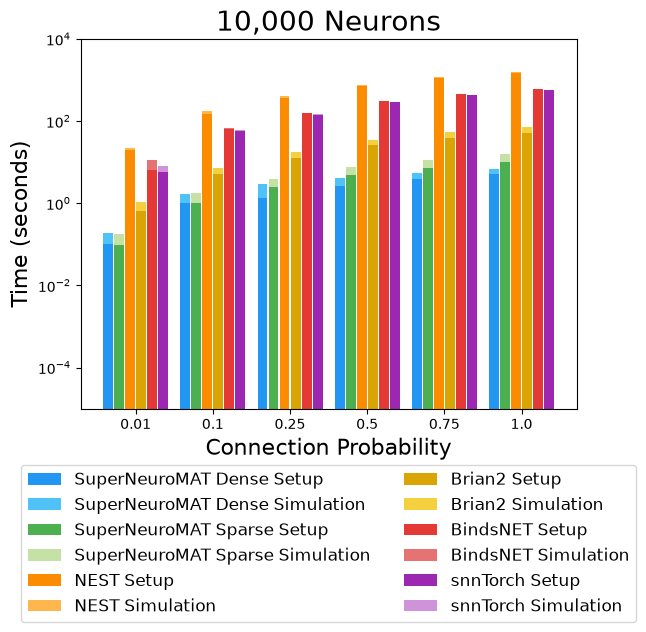}
        \caption{10,000 Neurons}
        \label{subfig:10000-neurons}
    \end{subfigure}
    \caption{Run time comparison for \snm sparse (green), \snm dense (blue), NEST (orange), Brian2 (yellow), BindsNet (red), and snnTorch (purple) for (a) 10 neurons; (b) 100 neurons; (c) 1,000 neurons; and (d) 10,000 neurons. The connection probability is shown on the X-axis, and the time in seconds is shown on the Y-axis.}
    \label{fig:performance-runtimes}
\end{figure}

\begin{figure}[t!]
    \centering
    \begin{subfigure}{0.49\textwidth}
        \centering
        \includegraphics[width=\textwidth]{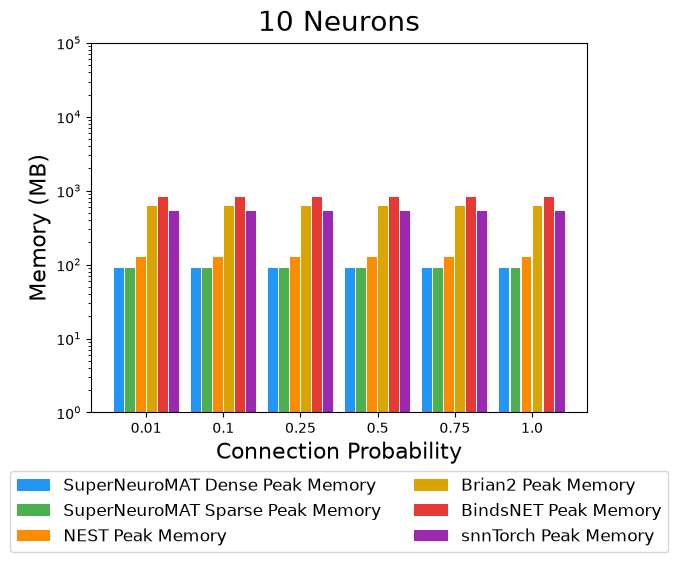}
        \caption{10 Neurons}
        \label{subfig:10-neurons}
    \end{subfigure}
    \begin{subfigure}{0.49\textwidth}
        \centering
        \includegraphics[width=\textwidth]{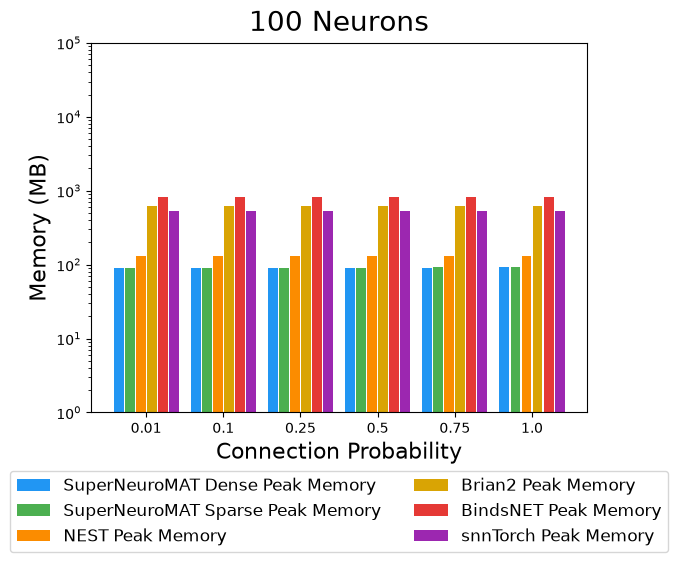}
        \caption{100 Neurons}
        \label{subfig:100-neurons}
    \end{subfigure}
    \\
    \begin{subfigure}{0.49\textwidth}
        \centering
        \includegraphics[width=\textwidth]{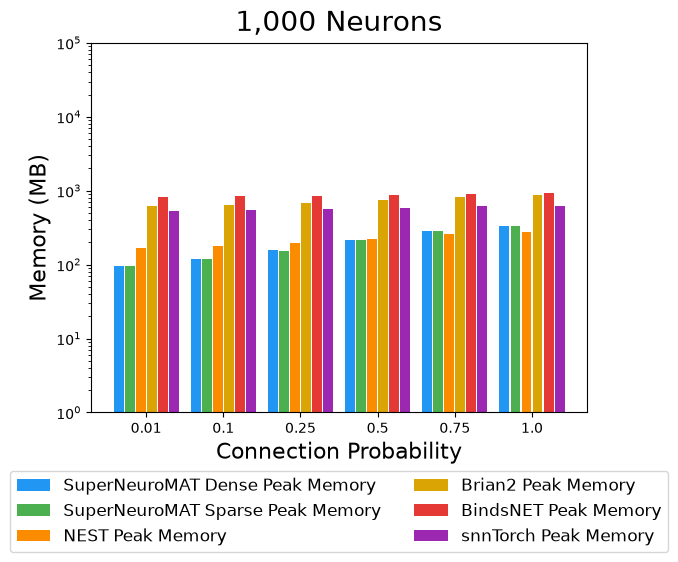}
        \caption{1,000 Neurons}
        \label{subfig:1000-neurons}
    \end{subfigure}
    \begin{subfigure}{0.49\textwidth}
        \centering
        \includegraphics[width=\textwidth]{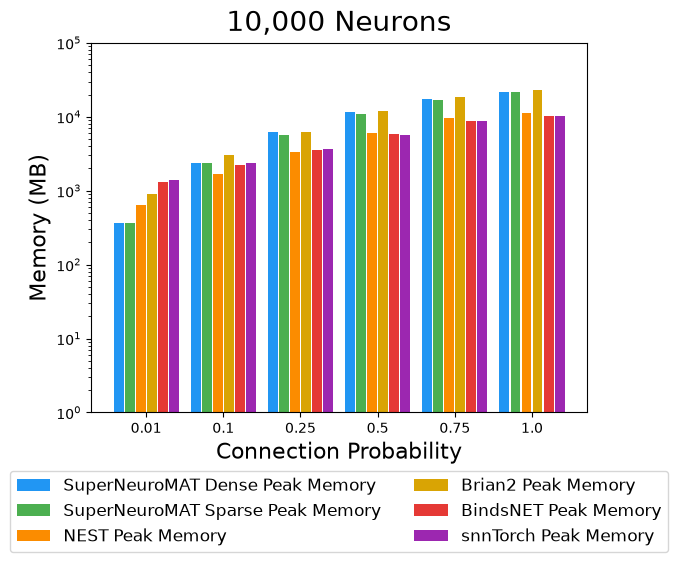}
        \caption{10,000 Neurons}
        \label{subfig:10000-neurons}
    \end{subfigure}
    \caption{Peak memory comparison for \snm sparse (green), \snm dense (blue), NEST (orange), Brian2 (yellow), BindsNet (red), and snnTorch (purple) for (a) 10 neurons; (b) 100 neurons; (c) 1,000 neurons; and (d) 10,000 neurons. The connection probability is shown on the X-axis, and the time in seconds is shown on the Y-axis.}
    \label{fig:performance-memory}
\end{figure}

We now evaluate the computational efficiency of both sparse and dense modes of \snm against four widely used SNN simulation frameworks: NEST, Brian2, BindsNET, and snnTorch.
Rather than relying on a specific application-derived network, the comparison uses synthetic Erd\H{o}s--R\'enyi random graphs spanning four network sizes (10, 100, 1,000, and 10,000 neurons) and six connection probabilities (0.01, 0.1, 0.25, 0.5, 0.75, and 1.0), with 10 independently generated graphs for each combination of network size and connection probability.
Each graph is converted into an equivalent SNN in every framework, and all neurons receive external spikes at every fifth time step.

While each simulator operates differently, we ensured that the neuron and synapse models are matched as closely as possible within what is allowed under each framework's API. 
In each simulator, we ensured that the neuron model being simulated matches the default configuration of \snm's neuron and synapse: zero threshold, memoryless (infinite-leak) LIF neurons with instantaneous (``delta'') synapses carrying a unit delay.
This was implemented via NEST's \texttt{iaf\_psc\_delta}, a custom-scheduled Brian2 equation set, BindsNET's \texttt{LIFNodes} with near-zero decay, and snnTorch's \texttt{Leaky} neuron with zero decay.
By equalizing simulation semantics across frameworks, differences in measured runtime and memory can be attributed to implementation efficiency rather than to differences in model complexity.
For each run, setup time (data-structure preparation) and simulation time (100 time steps of execution) were recorded separately, along with peak resident memory usage, and \snm was evaluated in both its dense (NumPy) and sparse (SciPy sparse matrix) modes.

Figure \ref{fig:performance-runtimes} shows the comparison of run times in seconds along the Y-axis as a function of increasing connection probability along the X-axis for four network sizes.
Across the tested range, \snm outperforms all four simulators, and its performance in relation to other simulators improves significantly as the network size increases.
At 1,000 neurons (averaged across all connection probabilities), \snm's dense mode completes in 0.052~s, versus 0.56~s for Brian2, 2.52~s for snnTorch, 2.81~s for BindsNET, and 7.05~s for NEST---a $10$--$135\times$ speedup.
At 10,000 neurons under full connectivity (the most computationally demanding configuration tested), \snm dense mode requires 7.18~s, compared with 70.0~s for Brian2 (9.8$\times$), 578.7~s for snnTorch (81$\times$), 613.3~s for BindsNET (85$\times$), and 1,545.6~s for NEST (215$\times$).
Notably, at the smallest network size (10 neurons) NEST is marginally faster than \snm, reflecting its low fixed per-run overhead.
However, this crossover disappears by 100 neurons, since NEST's per-neuron and per-edge object-creation calls scale poorly with network size, and \snm's advantage grows monotonically thereafter.

Figure \ref{fig:performance-memory} shows the comparison of peak resident memory in MB along the Y-axis as a function of increasing connection probability along the X-axis for four network sizes.
\snm maintains a consistently smaller memory footprint.
At small scale (10--100 neurons), its peak memory usage is negligible (well under 1~MB), whereas the other frameworks carry substantial fixed overhead---largely from their underlying runtime systems (PyTorch for BindsNET and snnTorch, the NEST kernel, and Brian2's code-generation backend)---ranging from roughly 130~MB (NEST) to over 800~MB (BindsNET).
At the largest scale tested (10,000 neurons, full connectivity), \snm consumes approximately the same peak memory as Brian2 and is slightly higher than the remaining three simulators.
However, the 2--3 orders of magnitude speedup obtained in the run times shown in Figure \ref{fig:performance-runtimes} by \snm as compared to NEST, BindsNET, and snnTorch, more than compensates for the slightly higher peak memory consumption.
Note that the memory values presented in Figure \ref{fig:performance-memory} include the memory consumed by the Erd\H{o}s--R\'enyi graphs, which were created using the \texttt{networkx} library in Python.

\subsection{Benchmark Examples}
\label{sub:benchmarks}

We provide six examples of how \snm can be used in a variety of different settings.
We focus on two conventional machine learning examples: the Digits dataset and citation networks datasets (Cora, Citeseer, and Pubmed).
Next, we delve into two neuromorphic ML datasets: the N-CARS dataset and the ASL-DVS dataset. 
Finally, we look into two non-ML examples: the shortest path algorithm and the arithmetic primitives.

\subsubsection{Digits}

\begin{figure}[t!]
    \centering
    \includegraphics[width=\textwidth]{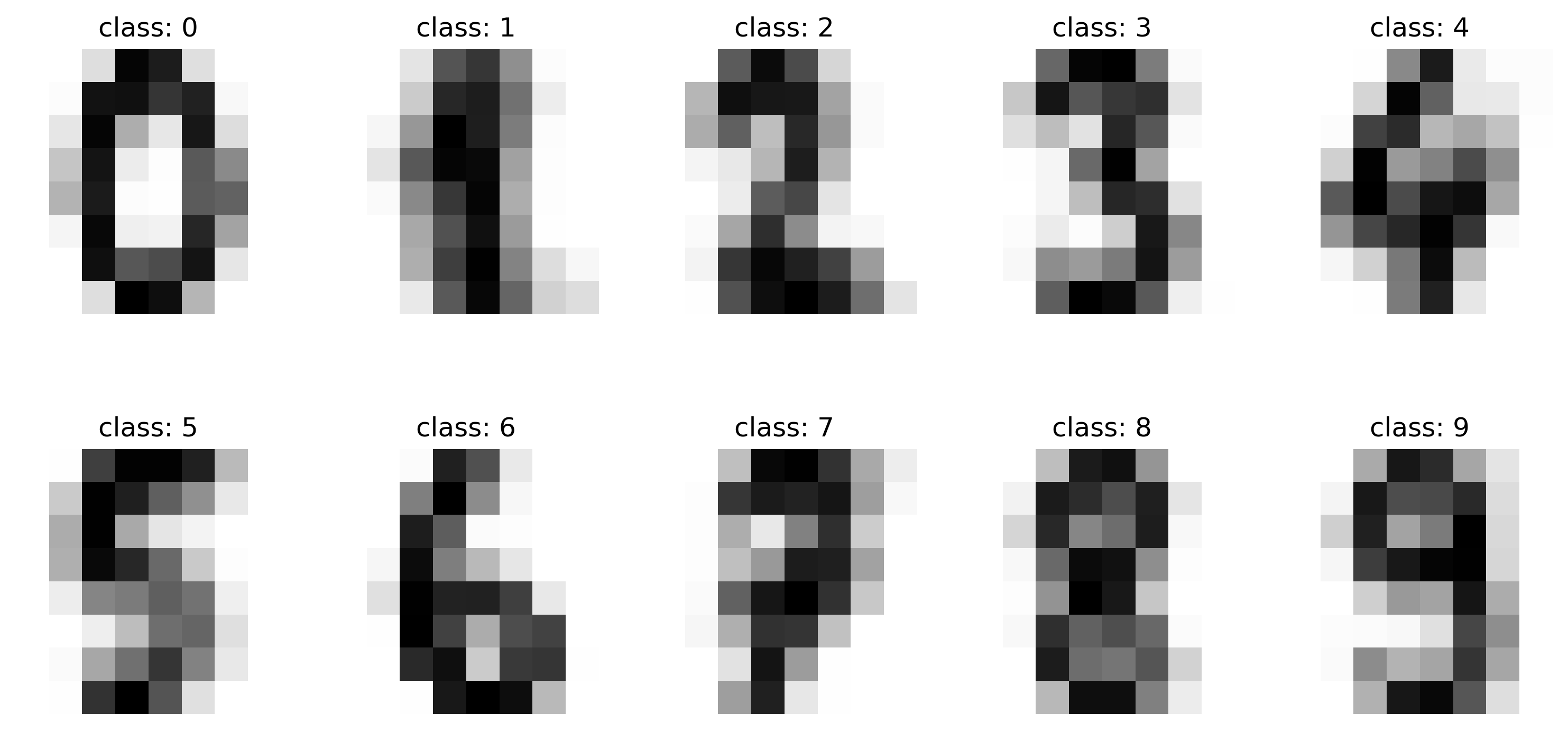}
    \caption{Learned weights for the Digits dataset using a two-layer SNN in \snm.}
    \label{fig:digits-weights}
\end{figure}

\begin{figure}[t!]
    \centering
    \includegraphics[width=0.5\textwidth]{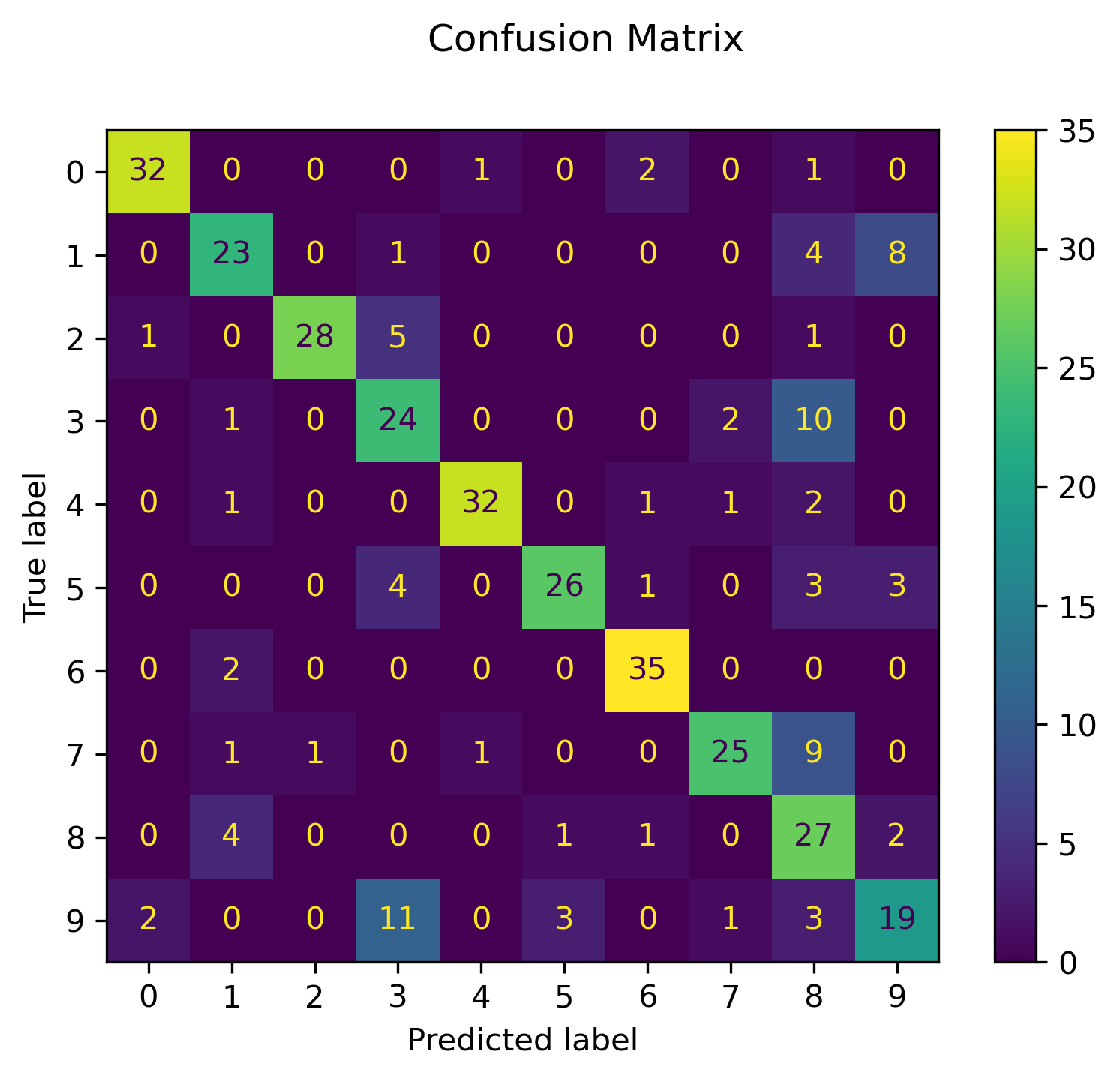}
    \caption{Confusion matrix for the Digits dataset using a two-layer SNN in \snm.}
    \label{fig:digits-confusion}
\end{figure}

We use \snm to implement a two-layer SNN for the Digits dataset, which is available in the Scikit-learn library in Python. 
The Digits dataset is a benchmark handwritten digit classification dataset containing $1{,}797$ grayscale images of decimal digits ($0$--$9$), with one class label per image.
Each image has a spatial size $8\times 8$, so every sample is represented by $64$ pixel features after flattening.
Pixel intensities are quantized integer values in the range $[0,15]$, corresponding to 4-bit grayscale levels.
We split the data into train and test subsets using a $80/20$ partition without shuffling, yielding approximately $1,438$ training samples and $359$ test samples.

Our SNN contained 64 input neurons for the 8$\times$8 pixels and 10 output neurons for the ten digit classes.
We encoded the pixel intensities as spike amplitudes and added input spikes to the correct output neuron during training to facilitate STDP-based synaptic updates.
Figure \ref{fig:digits-weights} shows the learned synaptic weights after the STDP operations.
After training, the learned weight matrix was used for inference with STDP disabled.
Predictions were decoded by selecting outputs with the highest spike counts and then resolving ties using earliest spike times.
We obtained $93\%$ one-versus-all accuracy, which is at par with the state-of-the-art for such a two-layer SNN.
Figure \ref{fig:digits-confusion} shows the confusion matrix for this dataset.
The Digits example demonstrates that a compact two-layer SNN can be successfully trained in \snm using built-in STDP.

\subsubsection{Citation Networks}

We used \snm for node classification on three citation network datasets (Cora, Citeseer, and Pubmed). 
These are well-established benchmarks for evaluating graph learning algorithms. In these datasets, each node corresponds to a scientific publication, directed edges denote citation relationships, and every publication belongs to one of several research topics ~\cite{sen2008collective,namata2012query}. Conventional approaches formulate this problem using graph neural networks (GNNs), where node attributes are iteratively aggregated through neighborhood message passing and optimized using backpropagation~\cite{kipf2017semi}. In contrast, we develop a neuromorphic formulation in \snm such that publications are represented as neurons and citation relationships as synapses.
This approach enables classification to emerge from event-driven spike propagation and local synaptic plasticity. 
This approach relies exclusively on graph topology and does not utilize node features, demonstrating that meaningful classification can be achieved through biologically plausible local learning operating directly on the graph structure.

Cora consists of 2,708 publications connected by 5,429 citation links and categorized into seven research topics. CiteSeer contains 3,327 publications, 4,732 citation links, and six topic classes, while PubMed comprises 19,717 publications connected by 44,338 citation links across three research areas. Standard Planetoid train/validation/test splits were adopted for all experiments~\cite{yang2016revisiting}. Each citation graph was mapped directly onto an SNN, with every publication represented by a neuron and every citation represented by a synapse. To suppress multiple spikes at a given neuron arising from cycles within the citation graph, neurons were assigned a sufficiently long refractory period such that each neuron emitted at most one spike during a simulation. A dedicated output neuron was allocated to each topic class, and every publication neuron in the test set was connected to all topic neurons through trainable synapses initialized with small random weights. During classification, a spike injected into a test publication neuron propagated through the citation graph according to its connectivity. Synaptic weights between publication neurons and topic neurons were updated online using STDP learning. At the end of the simulation, the predicted class was assigned to the topic neuron whose incoming synapse from the test publication neuron attained the largest weight.

Table~\ref{tab:citation_results} summarizes the classification performance---these are state-of-the-art results for spiking graph neural network models that do not incorporate features. 
Although the proposed method does not employ node features, global message passing, or gradient-based optimization, it successfully extracts discriminative information directly from graph topology through local spike-based computation.
Our results demonstrate that structurally meaningful representations can emerge from purely local synaptic adaptation.
This highlights the flexibility of \snm for graph-structured learning tasks beyond conventional neuromorphic sensory benchmarks.


\begin{table}[t]
\centering
\caption{Node classification performance on citation network benchmarks using graph topology alone. No node features were used during learning or classification.}
\label{tab:citation_results}
\begin{tabular}{lcccc}
\toprule
Dataset & Nodes & Citation Links & Classes & Accuracy (\%)\\
\midrule
Cora      & 2,708  & 5,429  & 7 & 62.7 \\
CiteSeer  & 3,327  & 4,732  & 6 & 42.4 \\
PubMed    & 19,717 & 44,338 & 3 & 54.7 \\
\bottomrule
\end{tabular}
\end{table}


\subsubsection{N-CARS}
\label{sec:ncars_dataset}

\paragraph{Approach 1: Diehl--Cook-Inspired STDP with Neuron-Level Fine-Tuning}
We used \snm to evaluate unsupervised SNN approaches for car-versus-background classification on the event-based N-CARS dataset~\cite{sironi2018hats}. The first approach employed a Diehl--Cook-inspired architecture comprising an input layer with 1,080 input neurons to encode the event camera voxel grid, a 64-neuron excitatory population, and a 64-neuron inhibitory population implementing winner-take-all competition~\cite{diehl2015unsupervised}. Input-to-excitatory synapses were trained without class labels using STDP, after which each excitatory neuron was assigned to the class that produced its highest average training set spike response.

\begin{figure}[t!]
    \centering
    \includegraphics[width=\linewidth]{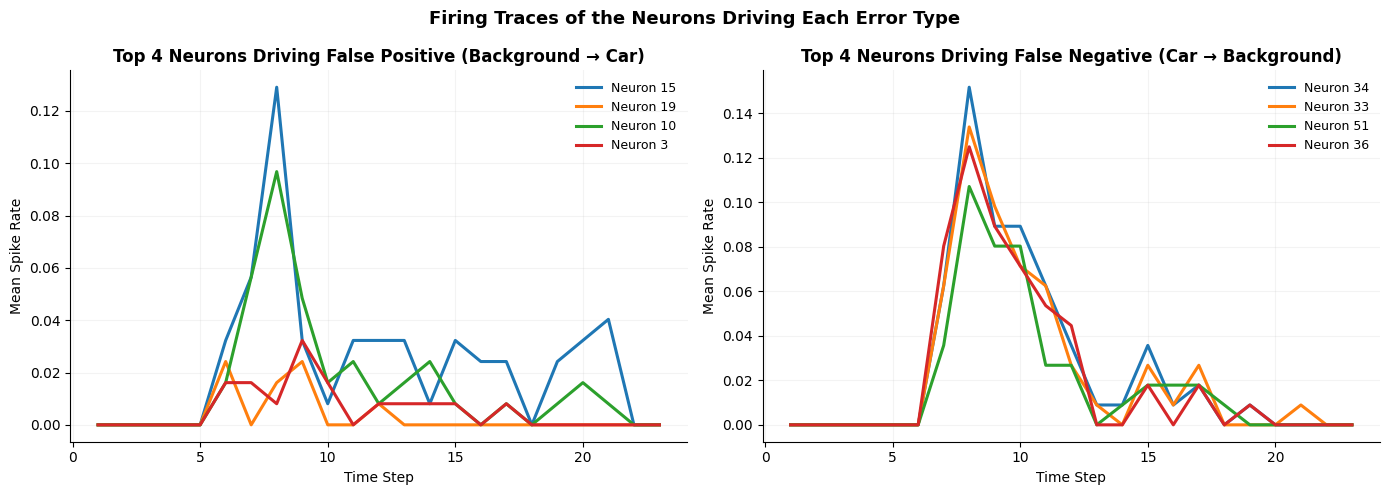}
    \caption{Mean spike activity of the excitatory neurons associated with false positive and false negative classifications across simulation time steps.}
    \label{fig:ncars_neuron_activity}
\end{figure}

\snm's spike train was then used to compare activity associated with true positive, true negative, false positive, and false negative predictions. 
As shown in Figure \ref{fig:ncars_neuron_activity}, two neurons exhibited firing rates at least 30\% higher during misclassifications, whereas six additional neurons displayed slowly decaying error-associated activity. The two neurons were therefore deactivated, and the leak values of the six remaining neurons were increased from 0.05 to 0.15. These targeted interventions improved the excitatory neuron accuracy by 4.3 percentage points, from 60.7\% to 65.0\%.
This example demonstrates how \snm can be used to interpret and enhance SNNs in an unsupervised learning setting.

\paragraph{Approach 2: Biologically Inspired Neuromorphic Feature Extraction}
We used \snm in a custom biologically inspired feature extraction pipeline to perform car-versus-background classification on the event-based N-CARS dataset~\cite{sironi2018hats}. Unlike conventional approaches that rely on established event representations, such as Histograms of Averaged Time Surfaces (HATS)~\cite{sironi2018hats}, Hierarchies of Time Surfaces (HOTS)~\cite{lagorce2017hots}, or related time-surface methods, the proposed approach learned a neuromorphic representation directly from the asynchronous event stream.

Events from the $100 \times 120$ pixel sensor were spatially mapped to a $10 \times 12$ lattice of ON- and OFF-polarity relay neurons and processed by a sparse 960-neuron SNN. The network incorporated center--surround filtering, fast and slow LIF traces, lateral inhibition, cross-polarity inhibition, and STDP at the excitatory trace-to-output synapses. The use of competitive lateral inhibition and unsupervised STDP builds on established SNN feature learning principles, including the competitive learning architecture introduced by Diehl and Cook~\cite{diehl2015unsupervised}.

\begin{figure*}[t!]
    \centering

    \begin{subfigure}[t]{0.48\textwidth}
        \centering
        \includegraphics[width=\linewidth]{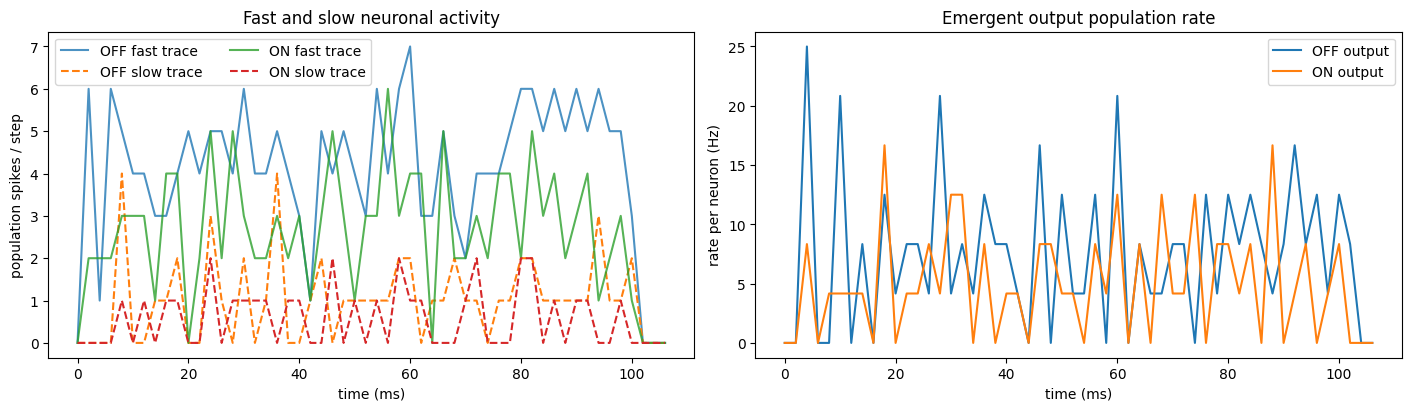}
        \caption{Population-level neural activity.}
        \label{fig:ncars_bio_neural_activity}
    \end{subfigure}
    \hfill
    \begin{subfigure}[t]{0.48\textwidth}
        \centering
        \includegraphics[width=\linewidth]{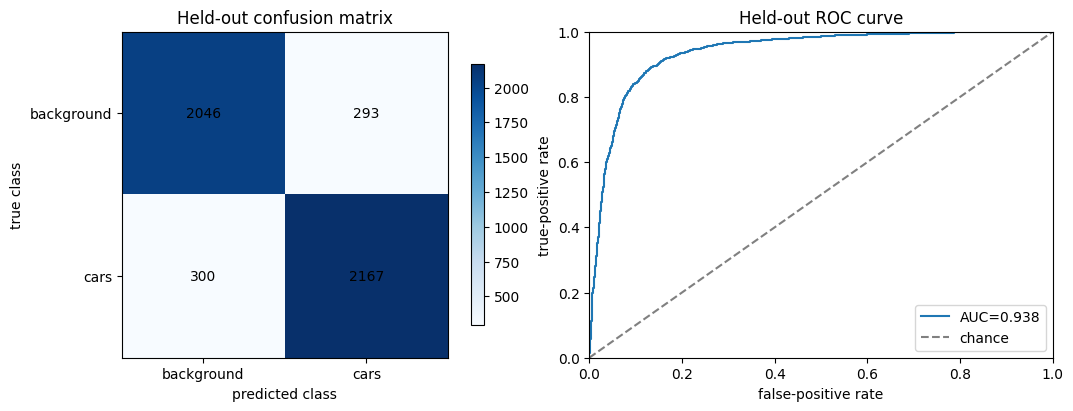}
        \caption{Logistic regression classification results.}
        \label{fig:ncars_bio_logistic_results}
    \end{subfigure}

    \caption{Analysis of the biologically inspired \snm representation on the N-CARS dataset.}
    \label{fig:ncars_bioinspired_analysis}
\end{figure*}

STDP training was performed in an unsupervised learning setting using the 80\% training dataset. After training, the SNN was frozen, and features derived from output spike counts, first-spike latencies, and population-level activity (Figure \ref{fig:ncars_bio_neural_activity}) were combined into a $2{,}880$-dimensional representation. This representation was then evaluated using both a regularized logistic regression classifier and a nonlinear multilayer perceptron with a $2{,}880 \rightarrow 128 \rightarrow 64 \rightarrow 2$ architecture. 
On the untouched 20\% test partition, comprising 4,806 recordings, the regularized logistic regression achieved 87.7\% accuracy, an F1 score of 88.0\%, and a receiver operating characteristic area under the curve (ROC-AUC) of 0.938 (Figure \ref{fig:ncars_bio_logistic_results}). 
When the same \snm representation was paired with the nonlinear MLP, performance improved to 92.4\% accuracy, an F1 score of 92.7\%, and a ROC-AUC of 0.975, demonstrating that the learned neuromorphic features were particularly effective when combined with a nonlinear classifier.

\subsubsection{ASL-DVS}
\label{sec:asldvs_dataset}

\begin{figure}[t!]
    \centering
    \includegraphics[width=\linewidth]{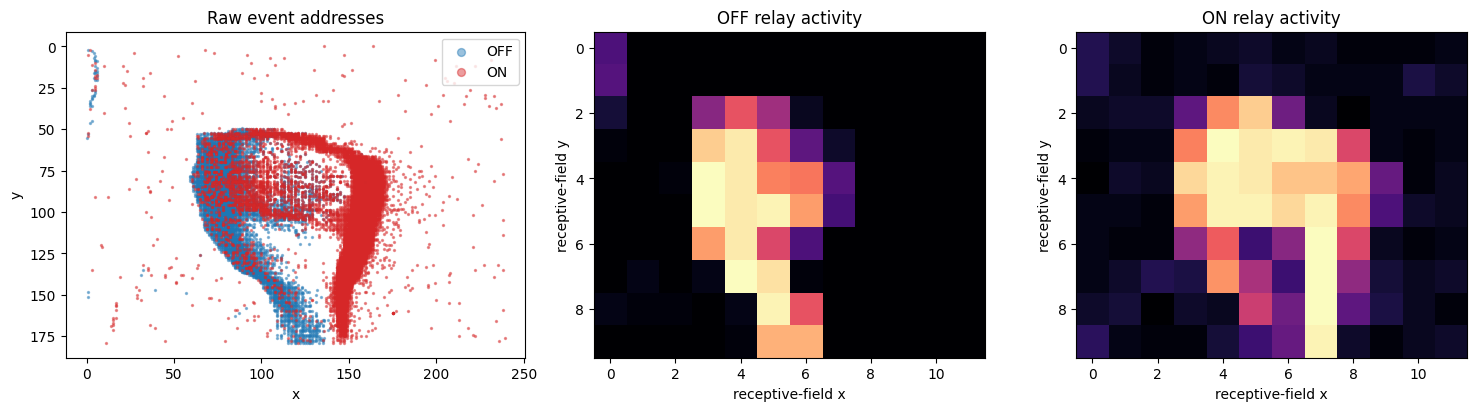}
    \caption{ASL-DVS letter classification results.}
    \label{fig:asldvs_results}
\end{figure}

We further evaluated the proposed \snm feature-extraction approach on the ASL-DVS dataset~\cite{bi2019graph}. ASL-DVS contains 100,800 event camera recordings representing 24 American Sign Language letters from A--Y, excluding J and Z. Each class contains 4,200 recordings captured at a spatial resolution of $240 \times 180$ pixels. 
An illustrative image is shown in Figure \ref{fig:asldvs_results}.
A stratified 80/20 partition was used in the present study, producing 80,640 training recordings and 20,160 held-out test recordings, with 840 test samples per class.

\snm was used to construct and simulate the SNN and perform STDP learning. Events were mapped to separate ON- and OFF-polarity pathways within a $10 \times 12$ receptive field lattice and processed using a $2\,\mathrm{ms}$ simulation time step. The resulting 960-neuron network incorporated center--surround filtering, fast- and slow-memory LIF neurons, lateral inhibition, and cross-polarity competition. STDP was restricted to the 480 excitatory trace-to-output synapses, which were trained without class labels using only the training partition.

Following unsupervised STDP training, the network parameters were frozen, and each recording was transformed into a 1,548-dimensional feature representation composed of output spike counts, first-spike latencies, and population-level activity measurements. The resulting features were standardized and supplied to a regularized logistic regression classifier.
On the test set, the SNN--STDP pipeline achieved 85.3\% accuracy, with a 95\% confidence interval of 84.8\%--85.7\%. Macro-averaged precision, recall, and F1 score were 85.3\%, 85.3\%, and 85.2\%, respectively, while the multiclass receiver operating characteristic area under the curve (ROC-AUC) reached 0.992. Classification performance was strongest for the G, P, Q, and H signs, whereas M, S, and T were among the most difficult classes to distinguish.

\subsubsection{Shortest Path Algorithm}

\begin{figure}[t!]
    \centering
    \includegraphics[width=\linewidth]{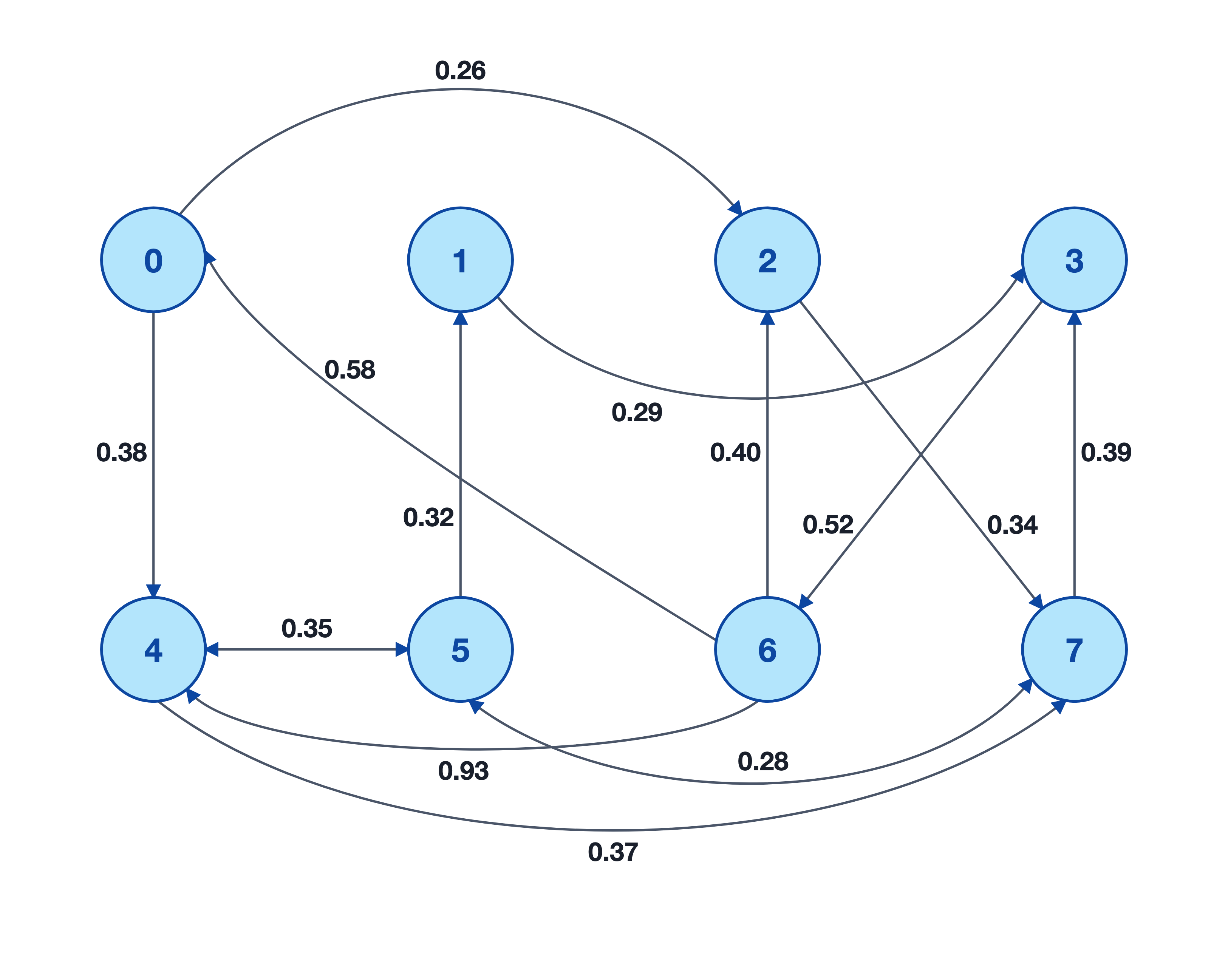}
    \caption{The graph containing eight nodes and fifteen edges used for comparing the neuromorphic shortest path algorithm implemented in \snm with Dijkstra's algorithm.}
    \label{fig:shortest-path-graph}
\end{figure}

\begin{table}[t!]
    \centering
    \caption{Shortest paths obtained from each source node to each destination node. These were obtained by implementing the neuromorphic shortest path algorithm in \snm. When scaled down by 100, these shortest paths match the shortest paths obtained using the Dijkstra's shortest path algorithm.}
    \label{tab:shortest-path-spike-times}
    \begin{tabular}{r r r r r r r r r}
        \toprule
        Source  & \multicolumn{8}{c}{Destination} \\
        \cmidrule{2-9}
                & 0 & 1 & 2 & 3 & 4 & 5 & 6 & 7 \\
        \midrule
        0 &   0 & 105 &  26 &  99 &  38 &  73 & 151 &  60 \\
        1 & 137 &   0 & 120 &  28 & 173 & 182 &  80 & 154 \\ 
        2 & 182 &  94 &   0 &  73 &  97 &  62 & 125 &  34 \\ 
        3 & 109 & 186 &  92 &   0 & 145 & 154 &  52 & 126 \\
        \midrule
        4 & 185 &  67 & 168 &  76 &   0 &  35 & 128 &  37 \\
        5 & 169 &  32 & 152 &  60 &  35 &   0 & 112 &  28 \\
        6 &  57 & 134 &  40 & 113 &  93 & 102 &   0 &  74 \\
        7 & 148 &  60 & 131 &  39 &  63 &  28 &  91 &   0 \\
        \bottomrule
    \end{tabular}
\end{table}

To evaluate \snm's ability to solve graph problems natively through spike timing, we implemented the neuromorphic shortest path algorithm \cite{kay2020neuromorphic} for a directed, weighted graph of 8 nodes and 15 edges as shown in Figure \ref{fig:shortest-path-graph} \cite{sedgewick_wayne_dijkstrasp}. 
Each node in the graph is represented by a neuron with a threshold of zero and a large refractory period such that it spikes exactly once during the course of the simulation. 
Each directed edge is represented by a synapse whose delay is proportional to the corresponding edge weight scaled by a factor of 100 to preserve two decimal digits of precision. 
Under this encoding, the arrival time of a spike at a given neuron corresponds to the cumulative weight of the path it traveled. 
A single input spike was injected into a given source node at time zero, and the network was simulated for 300 time steps.
This was repeated for all eight source nodes.

The spike times recorded for each neuron were decoded back into path distances by dividing by 100. 
The resulting values were compared against shortest distances computed using Dijkstra's algorithm. 
For all pairs of source and destination nodes, the neuromorphic shortest distances matched those from Dijkstra's algorithm. 
These are shown in Table \ref{tab:shortest-path-spike-times}. 
For instance, for the source neuron 0, neuron 2 spiked at time step $26$, corresponding to a distance of $0.26$, and neuron 6 spiked at time step $151$, corresponding to a distance of $1.51$. 
These results demonstrate the usability of \snm in solving the shortest path problem without requiring explicit priority queues or relaxation steps as required in Dijkstra's algorithm.


\subsubsection{Arithmetic Primitives}

\begin{figure}[t!]
    \centering
    \includegraphics[width=\linewidth]{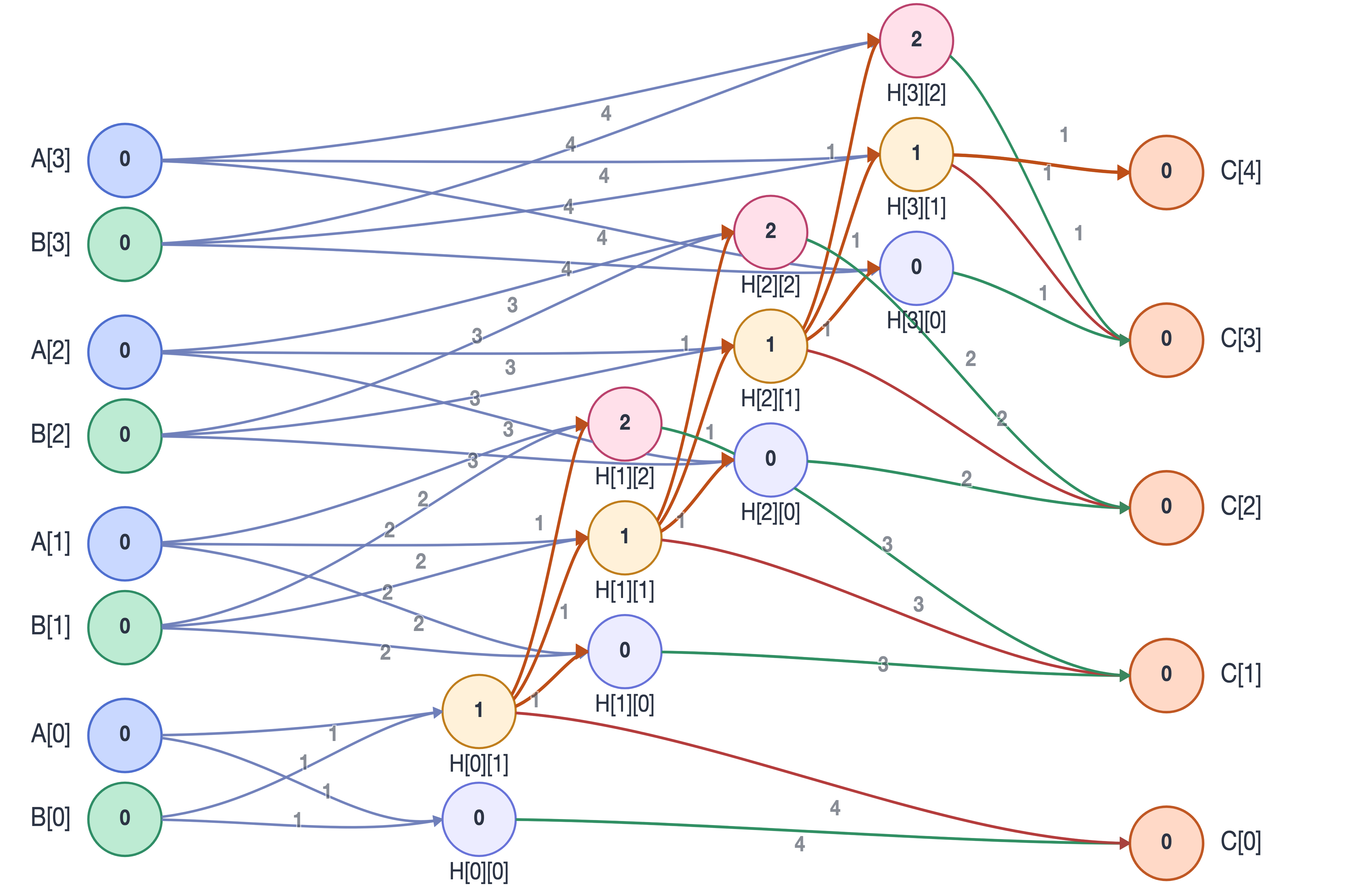}
    \caption{4-bit spiking adder. Takes two 4-bit operands ($A$ and $B$) encoded using binary encoding as inputs, performs the addition operation through the hidden neurons ($H$) and computes a 5-bit output ($C$).}
    \label{fig:adder-snn}
\end{figure}

To perform arithmetic operations using SNNs in \snm, we first encode each operand as a fixed-width binary integer, an $n$-bit integer across $n$ input neurons: neuron $i$ emits a spike if bit $i$ is 1. 
For addition, bit $i$ of each operand feeds a triad of neurons with thresholds $0$, $1$, and $2$ that reproduce the sum and carry logic of a full adder; the carry neuron at position $i$ is connected to the triad at position $i+1$ with a one time step delay, so that as the input spike wavefront propagates through the network, the carry ripples forward automatically. This results in a static, feed-forward ripple-carry adder whose control logic is the pattern of synaptic delays. 
Figure \ref{fig:adder-snn} shows a 4-bit adder as an illustration of this approach.
Because the SNN is purely feed-forward, its single instantiation can be pipelined such that successive operand pairs are injected one time step apart and are processed concurrently as they move through the network.
We validated our approach by building an 8-bit ripple-carry adder. Including the delay neurons, this adder SNN contained 293 neurons and 336 synapses. 
We streamed 100 randomly generated pairs of 8-bit unsigned integers through it and decoded the resulting spike trains back into decimal values. 
Comparing against conventional integer addition showed exact match for all 100 input pairs.

We also implemented an 8-bit multiplication SNN. 
Each partial product $A_i \wedge B_j$ was computed with a dedicated spiking neuron. 
The resulting eight shifted partial product rows were summed using a binary merge tree of delay-based ripple-carry adders rather than a single sequential accumulation: at each step, the two partial sums that are ready earliest are combined first, so most merges require no additional resynchronization delay and the SNN depth scales with $\log_2(N)$, where $N$ is the number of bits. 
Testing this SNN on 20 randomly generated pairs of 8-bit unsigned integers, the correct 16-bit product was recovered from the output spike times in all 20 cases, using 2,238 neurons and 2,802 synapses.


\section{Conclusion}
\label{sec:conclusion}

This paper introduced \snm, a fast, scalable, and open-source SNN simulator
that leverages a novel matrix-based approach to model the LIF operations, our simulator achieves exceptional computational efficiency on standard CPUs.
Both dense and sparse execution modes efficiently simulate up to 10,000 neurons under a dense implementation and approximately 100,000 neurons under a sparse implementation directly on personal computers. 

Our results indicate that \snm consistently outperforms four established SNN frameworks---NEST, Brian2, BindsNET, and snnTorch---in both total run time and peak resident memory across a wide range of connection probabilities and network sizes.
Furthermore, we showcased the versatility of \snm across six diverse examples. 
We applied \snm to conventional ML benchmarks (Digits and citation network datasets) as well as to neuromorphic event-based vision datasets (N-CARS and ASL-DVS). 
Finally, we validated its capability for non-ML, general-purpose computing by accurately implementing a neuromorphic shortest path algorithm and pipelined arithmetic primitives.

Our guiding rationale was to create a framework that is both highly performant and widely accessible; therefore, we developed it to be Python-based, user-friendly, and easily installable via PyPI. 
By lowering the barrier to entry and providing a powerful tool for SNN simulation and off-chip training, we intend for \snm to foster a collaborative global research environment and accelerate the broader development of SNNs and neuromorphic computing.

In the future, we would like to implement \snm in a distributed fashion to simulate billions of neurons and trillions of synapses.
Such an implementation will have to leverage both CPUs and GPUs.
Unlike desktops or laptops, this implementation would cater to exascale HPC systems such as Frontier, Lux, and Discovery.

\section*{Acknowledgements}
This material is based upon work supported by the U.S. Department of Energy, Office of Science, Office of Advanced Scientific Computing Research, under contract number DE-AC05-00OR22725. This manuscript has been co-authored by UT-Battelle, LLC under Contract No. DE-AC05-00OR22725 with the US Department of Energy. 
The United States Government retains and the publisher, by accepting the article for publication, acknowledges that the United States Government retains a non-exclusive, paid-up, irrevocable, world- wide license to publish or reproduce the published form of this manuscript or allow others to do so, for United States Government purposes. The Department of Energy will provide public access to these results of federally sponsored research in accordance with the DOE Public Access Plan (\url{http://energy.gov/downloads/doe-public-access-plan}).
The authors would like to thank John Baston III, Technical Editor at the Oak Ridge National Laboratory (ORNL), for reviewing and editing this manuscript.
The authors would like to thank Nicholas Quentin Haas, Software Developer at ORNL, for providing access to the virtual machine on which the computational efficiency results were obtained in this manuscript.

\bibliography{bib/references,bib/date}

\end{document}